\documentclass[11pt, a4paper, gr]{google}

\usepackage{hyperref}
\usepackage{url}
\usepackage{natbib}

\usepackage{booktabs}
\usepackage{multirow}
\usepackage{graphicx}
\usepackage{xspace}
\usepackage{cleveref}

\usepackage{wrapfig}

\usepackage[most]{tcolorbox}

\newcommand{\eg}{e.g.\xspace}

\newcommand{\ourmethod}{W\&L}

\usepackage[table]{xcolor}
\definecolor{lightgray}{gray}{0.9}

\uselogo{} 
\title{Watch and Learn: Learning to Use Computers from Online Videos}

\newtcolorbox{promptbox}[1][]{
    colback=blue!5!white,
    colframe=blue!75!black,
    fonttitle=\bfseries,
    title=Prompt,
    arc=4mm,
    boxrule=1pt,
    #1
}

\correspondingauthor{chanhee.luke@gmail.com, \{yiwensong, oriva, hamidpalangi\}@google.com}

\author[3+]{Chan Hee Song}
\author[1]{Yiwen Song}
\author[1]{Palash Goyal}
\author[3]{Yu Su}
\author[2*]{Oriana Riva}
\author[1*]{Hamid Palangi}
\author[1*]{Tomas Pfister}

\affil[1]{Google Cloud AI Research}
\affil[2]{Google DeepMind}
\affil[3]{The Ohio State University}

\begin{abstract}
Computer-using agents (CUAs) must plan task workflows across diverse and evolving applications, yet progress is limited by the lack of large-scale, high-quality training data. Existing datasets are narrow, static, and costly to annotate, while synthetic data often yields oversimplified or misaligned behaviors. 
We present \emph{Watch \& Learn} (\emph{\ourmethod}), a framework that converts readily available Internet videos of human computer use into executable UI trajectories at scale. 
Instead of directly generating actions or relying on handcrafted heuristics, we cast trajectory annotation as an inverse dynamics problem that predicts user actions from consecutive screen states, which simplifies learning and generalizes across domains. 
Through a task-aware retrieval and labeling pipeline, \ourmethod{} yields over 53K high-quality trajectories that enhance CUAs both as in-context exemplars and as supervised training data. 
On OSWorld, it consistently improves general-purpose and specialized CUAs, while on WindowsAgentArena it achieves state-of-the-art performance among 7B-scale models under the 15-step limit. 
These results show that web-scale human demonstration videos can serve as a practical and scalable foundation for advancing real-world CUAs. \\
\textbf{Project page:} \url{https://chanh.ee/wandl/}.
\end{abstract}

\begin{document}
\maketitle

\section{Introduction}
\label{sec:intro}

Computer-using agents (CUAs)~\citep{zheng2024seeact, Kil_2024_CVPR, uitars, gou2024uground, openai2025operator, yang2025ultracuafoundationmodelcomputer, wang2025opencua, jedi, Agent-S, Agent-S2, Agent-S3} are emerging as a powerful paradigm for automating software and web interactions. From everyday productivity tools to enterprise workflows, CUAs must both \emph{plan} multi-step task sequences that incorporate domain knowledge and \emph{ground} these plans into concrete UI actions across diverse, constantly changing interfaces. Despite rapid progress, their scalability remains constrained by the difficulty of obtaining large-scale, high-quality demonstrations of human-computer interaction~\cite{wang2025opencua}.

Collecting such trajectories remains extremely costly and time-consuming. For instance, OpenCUA’s AgentNet dataset~\citep{wang2025opencua}, which spans three operating systems and more than 200 applications and websites, required six months of dedicated annotation effort to label 22K tasks, with a total cost exceeding \$32{,}000 (about \$1.45 per task). Despite its impressive coverage, AgentNet underscores the scalability bottleneck of manual annotation: extending to the millions of trajectories needed for robust generalization would require substantial additional time and cost well over half a million dollars, even for well-funded labs. Such high costs and long turnaround times make it difficult to build datasets that capture the full diversity and complexity of real-world computer use, limiting progress in developing scalable computer-using agents.

\begin{figure}[tb]
  \centering
  \includegraphics[width=\linewidth]{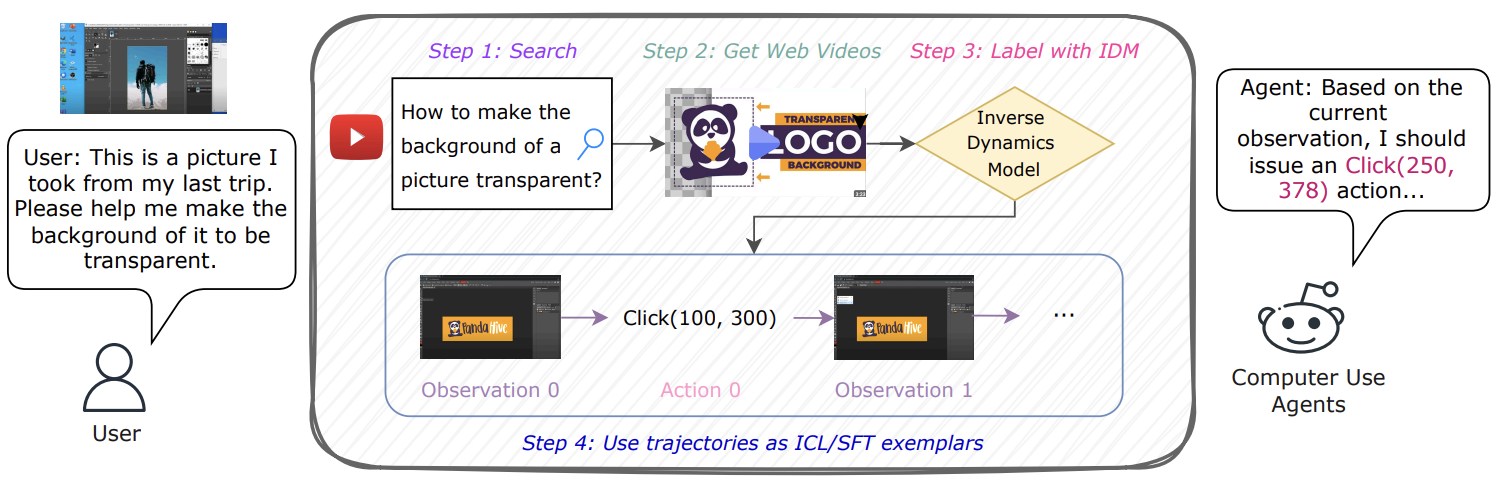}
  \caption{\ourmethod{} converts web-scale human demonstration videos into executable UI trajectories, providing scalable supervision and in-context exemplars for computer-using agents.}
  \label{fig:teaser}
\end{figure}

Meanwhile, the web already contains an enormous volume of human demonstration videos, including tutorials, screencasts, and walkthroughs, which naturally encode rich task workflows across applications. If we could reliably convert these videos into structured action trajectories, CUAs could learn directly from human expertise at unprecedented scale. However, existing trajectory generation methods have struggled to achieve this vision.

Prior work falls into three broad categories.  
\emph{Offline synthesis} reconstructs trajectories from recorded videos using multi-stage pipelines that couple multimodal large language models (MLLMs) with UI element detectors and transition parsers. Despite substantial engineering, systems such as MONDAY~\citep{jang2025_monday} and TongUI~\citep{zhang2025tongui} report modest accuracies (around 70\% for MONDAY) due to cascading errors in heuristic pipelines.  
\emph{Online synthesis} collects trajectories through random exploration and retrofits them with task instructions~\citep{murty2024bagel,osgenesis}, but these data are low in task complexity and expensive to gather.  
\emph{Hybrid approaches}, such as Explorer~\citep{pahuja-etal-2025-explorer}, mix proposal generation with online refinement, yet still depend on MLLMs for action grounding, retaining many of the same limitations.

To address these limitations, we introduce \textbf{Watch \& Learn (\ourmethod)}, a framework that formulates trajectory recovery as an \emph{inverse dynamics} prediction problem. Given two consecutive screen observations $(O_t, O_{t+1})$, the model predicts the intermediate action $a_t$ that produced the transition. This one-step formulation is substantially easier to learn than generating entire trajectories, avoids hand-crafted parsing heuristics, and generalizes across applications. In robotics, inverse dynamics modeling has long been used to infer actions from state transitions (\eg, VPT~\citep{baker2022videopretraining}, DreamGen~\citep{jang2025dreamgen}); here, we adapt this principle to computer use, showing it can recover accurate, executable actions from raw visual transitions.

We train an inverse dynamics model (IDM) on a large state-transition corpus containing more than 600K automatically labeled transitions from real-world web interactions. Each sample records $(O_t, a_t, O_{t+1})$, enabling direct supervision from interaction logs without manual annotation. We then pair this model with a task-aware retrieval module that locates relevant YouTube videos, either those specific to target benchmarks or general instructional content, and automatically converts them into structured UI trajectories. This pipeline yields 53{,}125 high-quality trajectories spanning a broad range of applications and task categories across three operating systems (Ubuntu, macOS, and Windows).

Beyond scalable data collection, \ourmethod{} reveals a dual role for such trajectories. They can be used not only for supervised fine-tuning, but also as \emph{in-context exemplars} at inference time, allowing CUAs to dynamically retrieve demonstrations as planning and grounding priors. This enables flexible integration with both open-source CUAs and general-purpose multimodal agents.

We evaluate \ourmethod{} on two challenging benchmarks, WindowsAgentArena~\citep{bonatti2024windowsarena} and OSWorld~\citep{xie2024osworld}, which require complex reasoning and precise UI grounding. On OSWorld, trajectories extracted from web videos yield consistent gains: in-context use improves general-purpose models and agentic frameworks by up to 3 percentage points, while supervised fine-tuning provides larger boosts for open-weight models (up to 11 points). On WindowsAgentArena, fine-tuning with our trajectories achieves leading performance among 7B-scale models under the 15-step evaluation limit, demonstrating that large-scale web demonstrations can translate effectively into high-quality training data. These benefits are obtained without any manual annotation, highlighting the potential of web-scale human workflows as a practical and scalable foundation for advancing CUAs toward real-world deployment. In summary, this work makes the following contributions:
\begin{itemize}
    \item We introduce a \textbf{inverse dynamics model} that recovers UI actions directly from visual transitions, removing the need for heuristic parsing or manual labeling.  
    \item We develop a \textbf{task-aware retrieval framework} that converts web-scale human demonstration videos into 53K executable UI trajectories.  
    \item We show that these trajectories \textbf{enhance CUAs} both as in-context exemplars and as supervised training data, achieving strong gains on OSWorld and WindowsAgentArena.
\end{itemize}

\section{Related Work}
\label{sec:related}

\subsection{Data Synthesis for Computer-Using Agents}

Human-curated UI datasets~\citep{mind2web,weblinx,aitw,android-control} remain limited in scale and diversity, constraining progress in CUAs. Recent research therefore explores synthesizing data through exploration, tutorials, or self-play.

Exploration-based synthesis methods such as BAGEL~\citep{murty2024bagel}, NNetNav~\citep{murty2025nnetscape}, Explorer~\citep{pahuja-etal-2025-explorer}, and OS-Genesis~\citep{osgenesis} let agents explore websites and retroactively label interactions with task instructions. This yields large-scale data but often introduces noise, since quality depends heavily on heuristics or multimodal LLM labeling.  

Tutorial-driven synthesis leverages online instructional resources. Synatra~\citep{ou2024synatra} and AgentTrek~\citep{xu2025agenttrek} convert textual tutorials into executable trajectories, while TongUI~\citep{zhang2025tongui} aggregates multimodal tutorials (text and screencast videos) into GUI data. These works show that web-scale content can broaden coverage across applications, though reliance on off-the-shelf LLMs often leads to brittleness and misalignment.

Self-improving agents integrate data synthesis directly into training. 
OpenWebVoyager~\citep{he-etal-2025-openwebvoyager} improves via online exploration and feedback, 
WebRL~\citep{qi2025webrl} creates new instructions from failures, 
SCA~\citep{qi2025webrl} generates and verifies code-based tasks, 
and ZeroGUI~\citep{yang2025zerogui} trains GUI agents entirely through automated task and reward generation. 
These systems enable continual learning without human data but often yield narrow or repetitive task distributions and require expensive iterative retraining.

Our approach, \emph{Watch \& Learn}, also mines web videos like TongUI but differs fundamentally in supervision. Rather than labeling videos with MLLMs, we train an inverse dynamics model that infers user actions directly from screen transitions. This produces accurate UI trajectories that strengthen both supervised training and in-context reasoning. By combining large-scale video mining with precise action labeling, our framework complements prior synthesis efforts and emphasizes the value of reliable video-based supervision for CUAs.

\subsection{In-context Learning for Agents}
In-Context Learning (ICL) has emerged as a pivotal test-time scaling paradigm for large language models, enabling them to adapt to new tasks without explicit parameter updates~\citep{dong2022survey}. This approach is particularly useful for enhancing LLM-powered agentic systems~\citep{su2025learn}.

Despite being generally helpful, the effectiveness of ICL is heavily influenced by the scale of the LLMs and the size of their context window, particularly for long-horizon, multi-step tasks. While including more ICL examples usually brings performance gains~\citep{agarwal2024many}, this method incurs significant computational overhead and latency with long demonstration trajectories. Therefore, efficiently selecting demonstration sequences~\citep{gupta2025leveraging} or abstracting them in high-level workflows~\citep{wang2024agentworkflowmemory,zheng2024synapse} has become a promising research direction. For computer-using agents, where tasks are often long and complex, one major challenge is the model's inability to plan effectively. Several pieces of work have leveraged ICL to address this specific problem~\citep{holt2025improving,zhao2025improving, Agent-S2, Agent-S, Agent-S3}. 

Another important direction is to develop data-centric
frameworks to adapt LLM agents to any given environments without human annotations \citep{su2025learn}. However, such methods require generating large amounts of synthetic data, and the potential for using publicly available web-scale video data as ICL examples still remains underexplored.

\begin{figure*}[tb]
  \centering
  \includegraphics[width=1\linewidth]{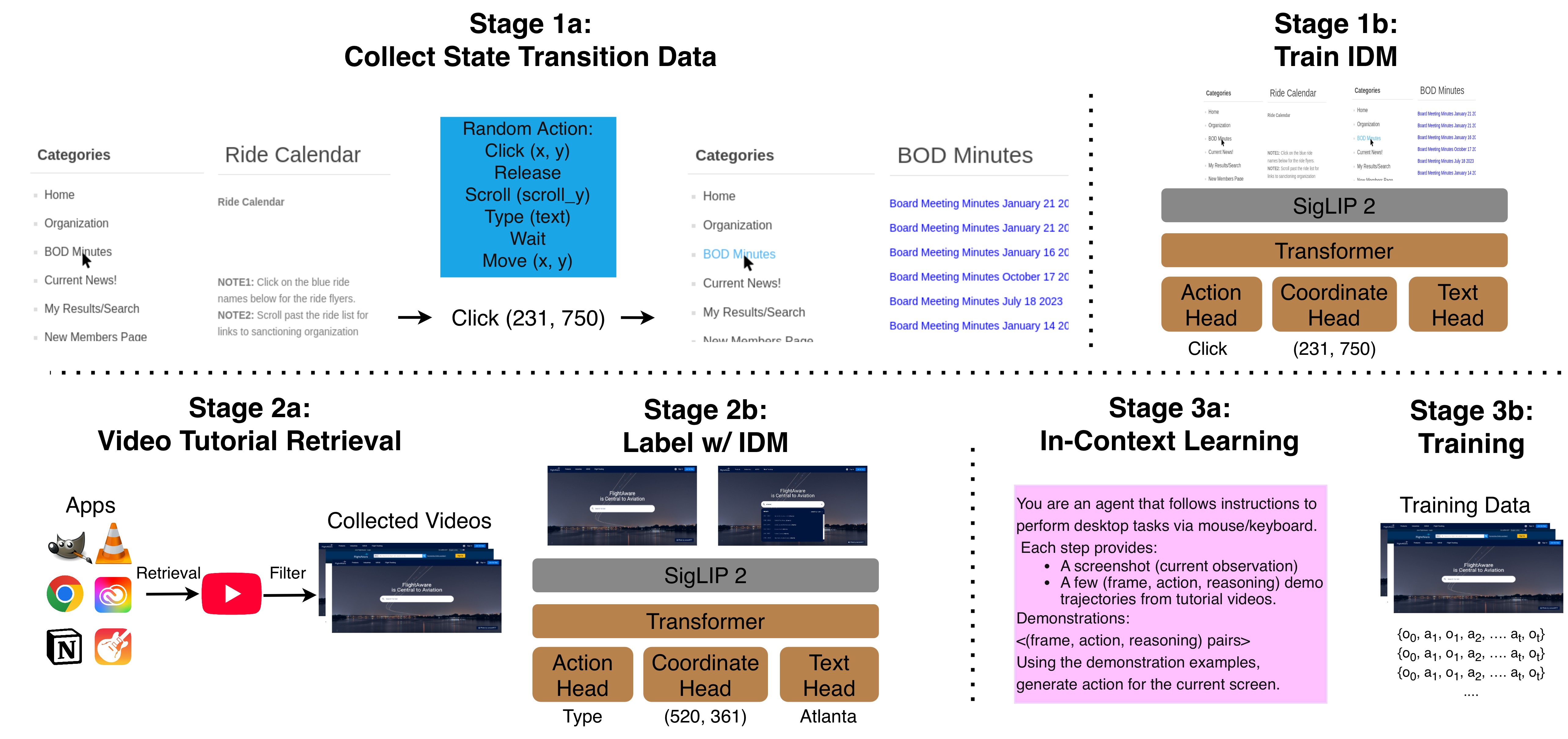} 
   \caption{\textbf{Method overview.} Our framework converts web-scale human demonstration videos into executable trajectories for CUAs. We first collect a large-scale state-transition dataset of screen observations and user actions, and train an inverse dynamics model (IDM) to recover actions from consecutive screenshots. This IDM is then applied to tutorial videos to extract step-by-step trajectories. A retrieval module selects task-relevant or general demonstrations, which are used in two ways: (i) as in-context exemplars that provide application-specific knowledge at inference time, and (ii) as supervised training data to improve open-source CUAs.
   }
   
  \label{fig:method}
\end{figure*}

\section{Method}
\label{sec:method}

Computer use agents must operate the user interface of many diverse and ever-changing applications where internal UI representations such as HTML or accessibility trees are often incomplete, inconsistent, or unavailable. To maximize generality and scalability, we focus on a \emph{vision-only} setting: models observe raw screen pixels and output structured user actions. This mirrors how humans interact with computers, by visually perceiving the interface and deciding where to click or what to type, while avoiding brittle dependencies on application-specific APIs or noisy UI representations.

At a high level, our framework works in three stages (see Figure~\ref{fig:method}). First, we construct a large-scale state-transition corpus from diverse computer interaction data and use it to train an inverse dynamics model (IDM), enabling the system to recover the underlying actions from consecutive screen observations. Second, we apply this IDM to web-scale tutorial videos, paired with a retrieval component that identifies either task-relevant videos (for inference-time use) or general tutorials (for training). This process automatically produces executable UI trajectories without manual labeling. 
Finally, we leverage these trajectories in two complementary ways: as \emph{in-context exemplars}, which provide CUAs with planning and grounding priors as well as application-specific knowledge at inference time; and as \emph{supervised training data}, which can be used to fine-tune models and improve their general knowledge.

\subsection{Inverse Dynamics Model}
\label{sec:method:idm}

A key component of our framework is an inverse dynamics model (IDM) that predicts the user action given two consecutive screen observations. Since large-scale state-transition data are scarce, we construct a new corpus by synthesizing interactions at scale.

\noindent \textbf{State-transition data.}
To obtain large-scale supervision, we built an automated pipeline that interacts with live web pages and records transitions. Inspired by WebDreamer~\citep{webdreamer}, we sample entry points from the March~2025 Common Crawl index and launch browsing sessions that execute actions drawn from the above schema. The sampling policy favors frequent actions (\eg, \texttt{click}) while ensuring sufficient diversity. This process yields roughly 600K $(O_t, a_t, O_{t+1})$ transitions. Further implementation details are provided in the Appendix.

\noindent \textbf{Action space.}
The IDM models six atomic primitives that together capture the majority of real-world computer interactions: \texttt{Click}, \texttt{Release}, \texttt{Scroll}, \texttt{Type}, \texttt{Wait}, and \texttt{Move}. Each action is parameterized by coordinates, text, or scalar values as appropriate. The \texttt{Click} primitive corresponds to a press event and includes normalized screen coordinates $(x, y)$, whereas \texttt{Release} denotes the corresponding button-up event and requires no positional arguments. This separation simplifies supervision and aligns with low-level GUI event semantics, where drag-and-drop gestures can be represented compositionally as a sequence of \texttt{Click} $\rightarrow$ \texttt{Move} $\rightarrow$ \texttt{Release}. Modeling these events as distinct atomic classes allows the IDM to predict a single action per transition while maintaining the flexibility to represent both short and prolonged clicks. Overall, this action space balances coverage and learnability, providing sufficient expressiveness for all interaction types required by OS-level benchmarks such as OSWorld and WindowsAgentArena.

\noindent \textbf{Model architecture.}
The IDM takes as input two consecutive observations $(O_t, O_{t+1})$ and outputs the action $a_t$ that caused the transition. Inspired by IDM architectures for robotics~\citep{baker2022videopretraining, jang2025dreamgen}, we adopt a vision-only architecture consisting of a SigLIP-2 vision encoder~\citep{tschannen2025siglip} followed by four Transformer~\citep{vaswani2017attention} layers. On top of this backbone, we attach three specialized prediction heads:
\begin{itemize}
    \item \textbf{Action classification head:} a categorical predictor over six supported primitives: \texttt{click}, \texttt{release}, \texttt{scroll}, \texttt{type}, \texttt{wait}, and \texttt{move}.
    \item \textbf{Coordinate head:} for location-based actions (\texttt{click}, \texttt{move}, and \texttt{type}), the model predicts normalized $(x, y)$ coordinates discretized into integers from $0$ to $999$. This formulation converts coordinate regression into a classification problem, which proved more stable during training.
    \item \textbf{Language head:} for text-entry actions, the model predicts the typed string autoregressively using a lightweight GPT-2 decoder~\citep{gpt2} conditioned on the visual features of $(O_t, O_{t+1})$. This design enables variable-length text generation while maintaining grounding in the observed state transition.
\end{itemize}
\texttt{Scroll}, \texttt{wait}, and \texttt{release} actions require no additional arguments; the model simply predicts their occurrence.

\noindent \textbf{Training.}
The IDM predicts three types of outputs depending on the action: (1) an action primitive, (2) discretized pixel coordinates $(x,y)$ for location-based actions, and (3) token sequences for text-entry.  
Given a transition $(O_t, O_{t+1})$, the model outputs a distribution over action primitives, optional coordinate distributions over $1000$ bins per axis (integers $0$--$999$), and an autoregressive distribution over text tokens.  
In practice, we discretize coordinates instead of regressing in continuous space, which stabilizes training and yields more accurate
click and typing locations.

We train the model with a multi-task objective that activates only the losses relevant to the ground-truth action. Concretely, we apply a classification loss for the primitive, a coordinate loss for location-based actions, and a language-modeling loss for text-entry.
Each head uses a standard cross-entropy term, and all losses are weighted equally unless otherwise specified. Full loss definitions are in the Appendix.

\noindent \textbf{Evaluation.}
We evaluate the IDM on a held-out subset of the state-transition corpus containing 100 transitions per action type. This evaluation set spans diverse real-world web interactions collected using the same automated pipeline described earlier. We assess performance along two dimensions: \emph{ActionType accuracy}, which measures the correctness of the predicted action primitive, and \emph{Action accuracy}, which additionally includes the precision of the predicted coordinates and generated text. As shown in Section~\ref{sec:exp:label-accuracy}, the IDM trained solely on state-transition supervision achieves strong action classification accuracy, outperforming off-the-shelf foundation models and confirming its effectiveness as the labeling backbone of our framework.

\subsection{Trajectory Data Generation from Videos}
\label{sec:method:videos}

Once the IDM is trained, we retrieve suitable tutorial videos and use the IDM to convert them into UI trajectories.

\noindent \textbf{Video retrieval.}
We build a retrieval framework that searches and downloads tutorial videos from large video platforms such as YouTube. The retrieval procedure differs depending on whether the goal is inference-time support or large-scale training data collection.  
\emph{Inference-time retrieval.} Given a task description and the target application, we form a natural language search query. To refine the query, we prompt Gemini~2.5 Flash\footnote{\scriptsize{\url{https://generativelanguage.googleapis.com/v1beta/models/gemini-2.5-flash:generateContent}}}~\citep{comanici2025gemini25} with both the task instruction and the initial screen, asking it to generate a more specific query. We then use the YouTube Search API to retrieve the top 15 videos. For example, a task instruction \texttt{"Can you increase the max volume of the video to the 200\% of the original volume in VLC?"} becomes the search query \texttt{"vlc increase max volume"}. Each retrieved video is paired with its title, which we treat as the candidate task description.  
\emph{Training-time retrieval.} To construct a broad training dataset, we curate a list of 69 applications spanning productivity, programming, design, screen editing, audio production, system utilities, and science/data domains (Table~\ref{tab:video_categories}). For each application, we prompt Gemini~2.5 Flash to generate plausible task queries and use them to search on video platforms, downloading the corresponding tutorial videos.

\noindent \textbf{Filtering.}
Retrieved videos vary widely in quality and content, so we first sample frames at 1 frame per second and discard unusable segments. 
Segments that are not screencasts (\eg, talking-head clips), are zoomed or cropped, or are blurred due to scene transitions are automatically removed by prompting Gemini~2.5 Flash. 
At inference time, we retain only the top 3 videos that pass this filter to reduce noise, whereas for training data collection we keep all qualified videos.

\noindent \textbf{Trajectory labeling.}
After filtering, we segment each video into a sequence of frames $\{O_0, O_1, \ldots\}$ and apply the IDM to every consecutive pair $(O_t, O_{t+1})$, predicting the intermediate action $a_t$ and assembling a trajectory $\tau = (O_0, a_0, O_1, a_1, \dots, O_{T}, a_{T}, O_{T+1})$. In this way, raw human demonstration videos are transformed into structured, executable trajectories without manual annotation. For inference-time usage, these trajectories are aligned with the task description and used as exemplars; for training-time usage, they are aggregated into a large corpus for supervised fine-tuning.

\begin{table}[tb]
\centering
\caption{Distribution of collected videos across 69 applications in 7 main categories.}
\begin{tabular}{lrr}
\toprule
Category & \# Apps & \# Videos \\
\midrule
Productivity     & 11 & 8,691 \\
Programming      & 12 & 12,829 \\
Design           &  9 & 7,948 \\
Screen Editing   &  8 & 7,808 \\
Audio Production &  8 & 5,206 \\
System Utilities & 11 & 4,601 \\
Science \& Data  & 10 & 6,042 \\
\midrule
\textbf{Total}   & \textbf{69} & \textbf{53,125} \\
\bottomrule
\end{tabular}
\label{tab:video_categories}
\end{table}

\begin{table*}[t]
\centering
\caption{
\textbf{Main results on OSWorld-Verified~\citep{xie2024osworld}} under the 50-step evaluation limit. 
IDM-labeled video exemplars from \ourmethod{} improve both general multimodal models (Gemini, Claude, o3, Jedi) via ICL and open-source CUAs via SFT, 
achieving up to $+11.1$ absolute gains over TongUI-labeled and baseline models.
}

\resizebox{1\textwidth}{!}{
\begin{tabular}{l l l l} 
\toprule
\textbf{Category} & \textbf{Base Model} & \textbf{Method} & \textbf{Success Rate (\%)} \\
\midrule
\rowcolor{lightgray}
\multicolumn{4}{c}{\textit{In-Context Learning}} \\
\hline

\multirow{7}{*}{General Models} & \multirow{2}{*}{Gemini 2.5 Flash~\citep{comanici2025gemini25}} & Base (w/o video) & 19.0 \\
 & & w/ \ourmethod{} labeled videos & \textbf{22.0 (+3.0)} \\
\cmidrule{2-4}
 & \multirow{3}{*}{OpenAI o3~\citep{openai2025o3o4mini}} & Base (w/o video) & 21.8 \\
 & & w/ TongUI labeled videos & 21.1 (-0.7) \\
 & & w/ \ourmethod{} labeled videos & \textbf{24.3 (+2.5)} \\
\cmidrule{2-4}
 & \multirow{3}{*}{Claude 4 Sonnet~\citep{anthropic2025claude}} & Base (w/o video) & 43.9 \\
   & & w/ TongUI labeled videos & 43.4 (-0.5) \\
 & & w/ \ourmethod{} labeled videos & \textbf{45.5 (+1.6)} \\
\midrule
\multirow{2}{*}{Agentic Framework} & \multirow{2}{*}{Jedi~\citep{jedi}} & Base (w/o video) & 50.6 \\
 & & w/ \ourmethod{} labeled videos & \textbf{52.8 (+2.2)} \\

\midrule 
\rowcolor{lightgray}
\multicolumn{4}{c}{\textit{Supervised Fine-Tuning}} \\
\hline

\multirow{7}{*}{Open-Source Models} & \multirow{3}{*}{Qwen 2.5VL 7B~\citep{qwen25vl}} & Base (No SFT) & 1.9 \\
 & & SFT w/ TongUI labeled & 5.4 (+3.5) \\
 & & SFT w/ \ourmethod{} (IDM labeled) & \textbf{13.0 (+11.1)} \\
\cmidrule{2-4}
 & \multirow{3}{*}{UI-TARS-1.5-7B~\citep{uitars}} & Base (No SFT) & 27.3 \\
 & & SFT w/ TongUI labeled & 23.8 (-3.5) \\
 & & SFT w/ \ourmethod{} (IDM labeled) & \textbf{31.1 (+3.8)} \\

\bottomrule
\end{tabular}
}
\label{tab:osworld_results}
\end{table*}

\begin{table}[t]
\centering
\caption{
\textbf{WindowsAgentArena results under the 15-step evaluation limit.}
Comparison of open-source 7B-sized CUAs.
\ourmethod{} achieves the highest success rate among 7B-sized models, surpassing both prior automatic labeling methods and zero-shot baselines.
}
\begin{tabular}{l l l}
\toprule
\textbf{Model} & \textbf{Setting / Training Data} & \textbf{Success Rate (\%)} \\
\midrule
\multirow{3}{*}{UI-TARS-1.5-7B~\citep{uitars}} 
 & Zero-shot (no SFT) & 18.1 \\
 & SFT w/ TongUI~\citep{zhang2025tongui} & 12.9 (-5.2) \\
 & SFT w/ \ourmethod{} (IDM-labeled) & \textbf{24.0 (+5.9)} \\
\midrule
OpenCUA-7B~\citep{wang2025opencua} & Base & 13.5 \\
UltraCUA-7B~\citep{yang2025ultracuafoundationmodelcomputer} & Base & 21.7 \\
\bottomrule
\end{tabular}
\label{tab:windows_results}
\end{table}

\subsection{Applications of Trajectories}
\label{sec:method:applications}

The extracted trajectories are utilized in two settings.
For in-context learning (ICL), each trajectory is converted into a set of \textit{(observation, action)} pairs augmented with concise natural-language reasoning traces generated by Gemini~2.5~Flash . These \textit{(observation, action, reasoning)} exemplars are formatted as few-shot demonstrations in the model context, enabling general-purpose agents to leverage planning and grounding priors without additional training.
For supervised fine-tuning (SFT), the labeled trajectories are aggregated into a large corpus of \textit{(state, action)} sequences and used to optimize multimodal LLMs with a standard sequence modeling objective. We fine-tune UI-TARS-1.5~\citep{uitars}, a specialized computer-using agent, and Qwen~2.5-VL~\citep{qwen25vl}, a general multimodal model, demonstrating that the proposed data serve as a scalable and transferable supervision source across diverse models.

\begin{figure*}[tb]
  \centering
  \includegraphics[width=\linewidth]{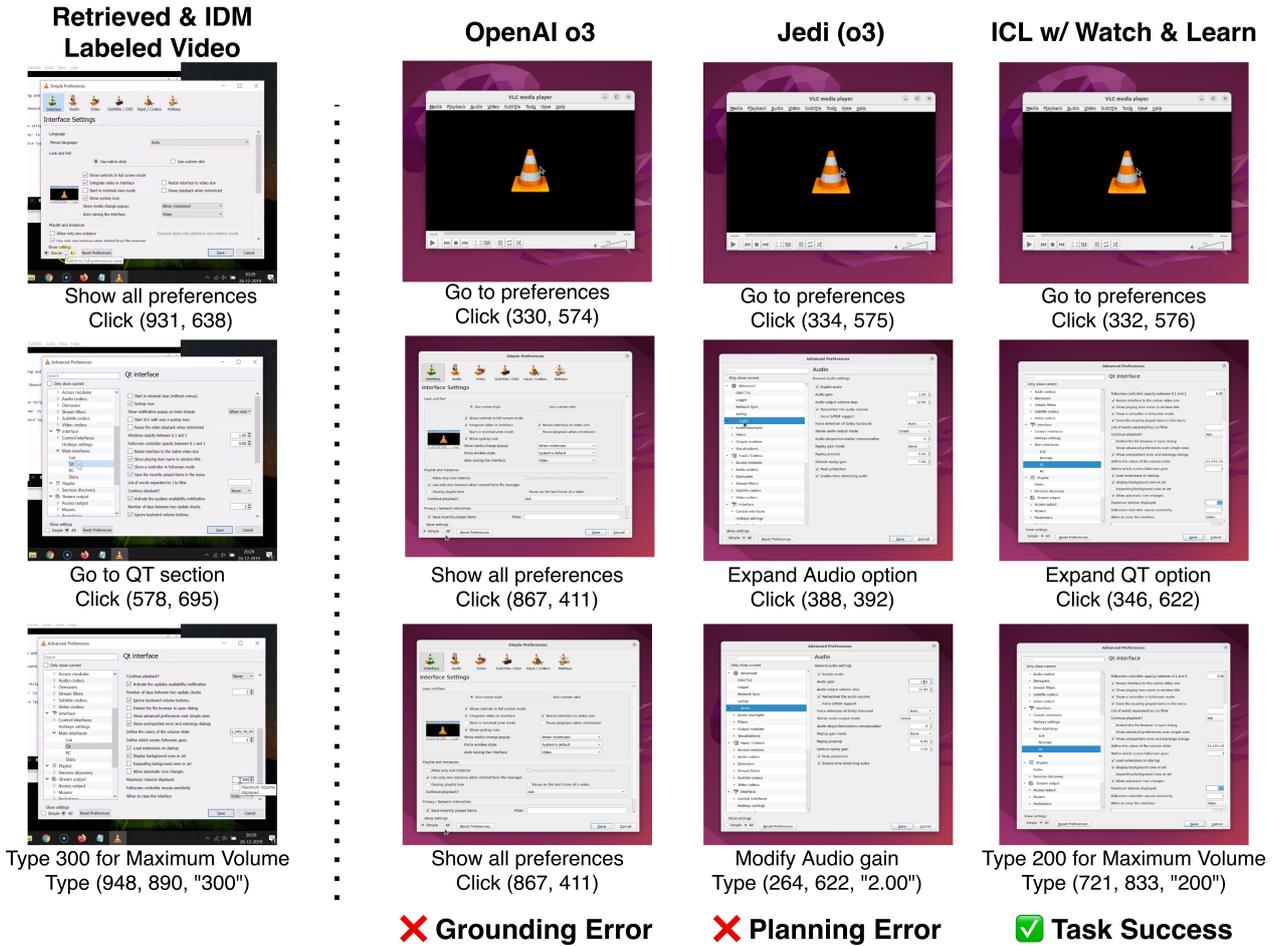} 
\caption{\textbf{Qualitative examples on OSWorld.} On the left, the video-derived trajectory that \ourmethod{} generates for the task. On the right: (i) the o3 agent makes a grounding error by selecting a wrong UI element; (ii) the Jedi (o3) agent makes a planning error by entering the wrong submenu without recovering; (iii) using the video-derived trajectory, \ourmethod{} agent completes the task successfully. Images are cropped for visibility, and the action coordinates correspond to the original full-resolution screenshots. More trajectory examples are in Appendix.}
  \label{fig:exp}
\end{figure*}

\section{Experiments and Analysis}
\label{sec:experiments}

\subsection{Experimental Setup}

Our experiments evaluate whether IDM-labeled trajectories improve interactive computer-using agents in two settings:  
(1) \emph{in-context learning (ICL)} for general-purpose multimodal models and agentic frameworks, and  
(2) \emph{supervised fine-tuning (SFT)} for open-source models.  
We also compare the resulting trajectories against TongUI~\cite{zhang2025tongui}, an existing automatic labeling pipeline.

All agents operate under a unified interface where each step provides only the current screen image and the history of executed actions.  
In the ICL setting, each model receives a fixed number of task-relevant trajectories retrieved from our video-derived dataset as exemplars.  
In the SFT setting, open-source CUAs are trained directly on the same IDM-labeled trajectories.

We evaluate on two complementary interactive computer-use benchmarks.  
OSWorld-Verified~\citep{xie2024osworld} provides full desktop and operating-system (Ubuntu) environments across productivity, programming, design, and system utilities, and we apply a 50-step interaction limit.  
To assess robustness across different interface conventions and OS-specific workflows, we additionally evaluate on WindowsAgentArena~\citep{bonatti2024windowsarena}, a benchmark of Windows tasks that follows a 15-step interaction limit.  
Together, these benchmarks allow us to test whether the benefits of IDM-labeled trajectories generalize across heterogeneous operating systems and UI distributions.

\subsection{Datasets}
\noindent \textbf{State-transition corpus.}
To train the IDM, we gather autonomous state transitions from real web interactions, resulting in over 600k $(O_t, a_t, O_{t+1})$ triples.

\noindent \textbf{Video-derived trajectories.}
After training the IDM, we apply it to a curated collection of 53{,}125 instructional YouTube videos, producing one trajectory per video and yielding 53{,}125 trajectories across 69 applications spanning productivity, programming, design, audio production, system utilities, and scientific/data domains (Table~\ref{tab:video_categories}). Additional details are provided in the Appendix.

\noindent \textbf{Baseline labeler.}
TongUI~\citep{zhang2025tongui} generates action annotations by prompting the UI-TARS-1.5-7B agent~\citep{uitars}, while our method applies the trained IDM to the same source videos.  
To ensure a fair comparison focused only on labeling quality, we use the identical set of tutorial videos retrieved by our framework and apply TongUI’s labeling pipeline, differing only in the model used to annotating actions.

\subsection{Models}

We evaluate three categories of models covering general multimodal model, agentic framework, and open-source CUAs.

\noindent \textbf{General-purpose multimodal models.}
Gemini~2.5 Flash~\citep{comanici2025gemini25}, OpenAI~o3~\citep{openai2025o3o4mini}, and Claude~4 Sonnet~\citep{anthropic2025claude} are evaluated in the ICL setting using retrieved IDM-labeled trajectories as exemplars.

\noindent \textbf{Agentic framework.}
We use Jedi~\citep{jedi}, a vision-only agentic framework for OSWorld that combines an MLLM planner (OpenAI~o3) with the Jedi-7B grounding model that maps natural language steps to executable UI actions.  
Its modular design allows straightforward integration of our ICL exemplars, so we evaluate Jedi both with and without retrieved IDM-labeled trajectories.

\noindent \textbf{Open-source CUAs.}
UI-TARS-1.5-7B~\citep{uitars} and Qwen~2.5-VL~7B~\citep{qwen25vl} are trained with supervised fine-tuning on our 53{,}125 video-derived trajectories, enabling a controlled comparison against TongUI-labeled data and the exemplar-based performance of multimodal models.

\subsection{Main Results}

Table~\ref{tab:osworld_results} summarizes our main results on OSWorld across both in-context learning and supervised fine-tuning. We observe consistent improvements across all model categories. For general-purpose multimodal models (Gemini~2.5 Flash, OpenAI o3, Claude~4 Sonnet), adding our \ourmethod{} exemplars improves performance by +1.6 to +3.0 points. This shows that trajectories distilled from web tutorials provide useful domain-specific priors that even strong foundation models can leverage at inference time. For the Jedi agentic framework~\citep{jedi}, which couples the o3 planner with Jedi-7B grounding model, \ourmethod{} yields a +2.2 point gain, demonstrating that our trajectories can complement structured planning pipelines by enriching them with exemplars that support both planning and grounding. For open-source CUAs, supervised fine-tuning on our 53k video-derived trajectories yields even larger gains. UI-TARS-7B improves by +3.8 points, while Qwen~2.5-VL sees the largest improvement, from 1.9 to 13.0 (+11.1). This larger jump is expected because Qwen is a general-purpose multimodal model not originally trained for computer use, so it benefits disproportionately from our dataset, which provides task-specific supervision that was previously missing. 

We further evaluate on WindowsAgentArena~\citep{bonatti2024windowsarena}, which focuses on Windows OS tasks under a standardized 15-step evaluation limit. As shown in Table~\ref{tab:windows_results}, \ourmethod{} achieves the state-of-the-art success rate among 7B-scale models with 15 step limit, surpassing both prior automatic labeling method and specialized CUA baselines. Specifically, fine-tuning UI-TARS-1.5-7B on IDM-labeled trajectories improves performance from 18.1 to 24.0~(+5.9), whereas training on TongUI-labeled data leads to a degradation in performance (12.9,~--5.2). These results confirm that IDM-based labeling produces more accurate and consistent supervision, translating to stronger downstream generalization. 

Overall, across both OSWorld and WindowsAgentArena, \ourmethod{} serves as a scalable and high-quality supervision signal that consistently enhances models and agentic frameworks in both in-context and fine-tuning settings.

\subsubsection{Effect of Label Accuracy on Performance}
\label{sec:exp:label-accuracy}

Action label accuracy is central to training CUAs: noisy annotations not only fail to help but can actively degrade performance. We first compare our dedicated IDM against Gemini~2.5 Flash and the TongUI labeling pipeline (based on UI-TARS-1.5-7B) on the held-out state-transition test set (Table~\ref{tab:idm_results}). Our IDM achieves the strongest results, substantially outperforming both baselines. TongUI offers modest gains over Gemini, especially for structured actions such as \texttt{scroll} and \texttt{click}, but still falls short of our IDM.

These differences in labeling accuracy directly translate into downstream performance. 
TongUI, although it follows the same prompt format, relies on labels generated by UI-TARS-1.5-7B, whose own performance is notably weaker on WindowsAgentArena (Table~\ref{tab:windows_results}) than on Ubuntu-based OSWorld (Table~\ref{tab:osworld_results}). 
As a result, labeling quality degrades more severely in the Windows setting, leading to noisier supervision. 
As shown in Table~\ref{tab:windows_results}, this effect is amplified on WindowsAgentArena, where fine-tuning with TongUI-labeled data causes a drop in success rate (a decrease of 5.2 points) compared with consistent gains from IDM-labeled trajectories. 
In contrast, our IDM-derived labels remain reliable across both Ubuntu-style (OSWorld) and Windows-style (WindowsAgentArena) interfaces, underscoring that accurate and platform-robust supervision is essential for effective action grounding and cross-OS generalization.

\subsubsection{Effect of Trajectories on In-Context Learning}

We next analyze the contribution of accurate video labeling to in-context learning (ICL). Our framework provides structured action annotations and natural language reasoning for each step. To isolate the effect of each component, we compare three variants: (i) consecutive frames only, (ii) frames paired with predicted actions, and (iii) frames with both actions and reasoning generated by Gemini~2.5 Flash.  

Ablations on OSWorld (Table~\ref{tab:icl_ablation}) show that adding action labels provides a substantial boost over using frames alone, and further gains are achieved when natural language reasoning is included. This pattern holds consistently across all tested models. Figure~\ref{fig:exp} illustrates these effects qualitatively: baseline agents without labeled trajectories often make grounding or planning errors, for example clicking incorrect UI elements or entering the wrong submenu, while the \ourmethod{}-enhanced agent successfully completes the same tasks by leveraging accurate video-derived exemplars.  
These observations confirm that labeled trajectories convey procedural and causal knowledge beyond visual context, improving both planning and grounding for complex workflows.

\begin{table}[t]
\centering
\caption{\textbf{Action labeling accuracy on the held-out test set.}
We evaluate six primitive action types with 100 transitions each.
\textit{ActionType Accuracy} reflects correct action classification, while \textit{Action Accuracy} also checks argument correctness.
\ourmethod's IDM outperforms Gemini and TongUI across all action types.}
\begin{tabular}{l c c c}
\toprule
ActionType & Gemini 2.5 Flash & TongUI & \ourmethod~IDM \\
\midrule
click(x, y) & 68\% & 72\% & \textbf{95\%} \\
release & 71\% & 67\% & \textbf{90\%} \\
scroll(scroll\_y) & 55\% & 75\% & \textbf{93\%} \\
type(text) & 77\% & 71\% & \textbf{86\%} \\
wait(500ms) & 92\% & 88\% & \textbf{97\%} \\
move(x, y) & 65\% & 61\% & \textbf{89\%} \\
\midrule
\textbf{ActionType Accuracy} & 81.5\% & 84.3\% & \textbf{95.8\%} \\
\textbf{Action Accuracy} & 70.5\% & 72.3\% & \textbf{91.7\%} \\
\bottomrule
\end{tabular}
\label{tab:idm_results}
\end{table}

\begin{table}[t]
\centering
\caption{\textbf{Ablation of action labeling and reasoning} within ICL exemplars on OSWorld. Structured trajectories lead to steady gains compared with raw frame exemplars.}
\begin{tabular}{lccc}
\toprule
 & Gemini 2.5 Flash & OpenAI o3 & Claude 4 Sonnet \\
\midrule
Baseline (no exemplars) & 19.0 & 21.8 & 43.9  \\
+ Frames                & 18.4 & 21.8 & 43.9  \\
+ Frames + Actions      & 20.1 & 23.0 & 44.4  \\
+ Frames + Actions + Reasoning & \textbf{22.0} & \textbf{24.3} & \textbf{45.5} \\
\bottomrule
\end{tabular}
\label{tab:icl_ablation}
\end{table}

\subsubsection{Effect of Video Retrieval on In-Context Learning}\label{sec:exp:retrieval-quality}

We further examine the role of retrieval quality by comparing our method against a random retrieval baseline using o3 (Table~\ref{tab:o3_random_retrieval}). Interestingly, random retrieval neither improves nor degrades performance relative to the base model. This suggests that, while carefully retrieved exemplars provide useful signal, even randomly selected exemplars do not introduce significant noise. A likely explanation is that the action labels themselves remain highly accurate regardless of retrieval quality, ensuring that the model is not misled by contradictory supervision. These results indicate that the main benefit of our method lies in providing \emph{targeted} exemplars that align closely with the task context. Retrieval quality therefore determines the strength of the positive effect, but poor retrieval does not actively harm performance when the underlying labels are still correct.

\begin{table}[t]
\centering
\caption{\textbf{Ablation of retrieval strategy for ICL on OSWorld.} Random retrieval gives minimal gains, while \textbf{task-relevant retrieval} in \ourmethod{} improves performance.}
\centering
\begin{tabular}{l c c c}
\toprule
 & o3 (base) & o3 + Random & o3 + \ourmethod \\
\midrule
ICL & 21.8 & 21.8 & \textbf{24.3 (+2.5)} \\
\bottomrule
\end{tabular}
\label{tab:o3_random_retrieval}
\end{table}

\section{Conclusion and Future Work}
\label{sec:conclusion}

We introduced \ourmethod, a framework that converts web-scale human demonstration videos into executable UI trajectories using a vision-only inverse dynamics model (IDM) and a task-aware video retrieval pipeline. With over 53k automatically labeled trajectories, \ourmethod{} improves both in-context learning and supervised fine-tuning across diverse agents, benefiting general-purpose multimodal models as well as open-source CUAs. Experiments on OSWorld and WindowsAgentArena confirm that IDM-labeled trajectories provide high-quality supervision that generalizes across operating systems and interface conventions.

Future work will focus on combining or segmenting tutorials to better capture long-horizon workflows and on exploring reinforcement learning using our trajectories as demonstrations, replay buffers, or priors for reward modeling. These directions aim to further bridge large-scale demonstrations with adaptive policy learning, bringing CUAs closer to robust real-world deployment.

\section*{Acknowledgments}

We thank the Google Cloud AI Research team for their valuable feedback, with special thanks to Yale Song for his constructive input. This work was supported in part by the National Science Foundation under NSF CAREER Award \#2443149 and by an Alfred P. Sloan Foundation Fellowship.

\bibliography{arxiv_google/bibfile}
\bibliographystyle{abbrvnat}

\clearpage

\appendix

\section*{Appendices}
\label{sec:Appendix}

In this supplementary material, we present additional details and clarifications that are omitted in the main text due to space constraints.
\begin{itemize}
    \item \autoref{supp:limitations} Limitations.
    \item \autoref{supp:dataset} Dataset Details.
    \item \autoref{supp:implementation} Implementation Details.
    \item \autoref{supp:idm_arch} Inverse Dynamics Model Architecture.
    \item \autoref{supp:idm_training} Inverse Dynamics Model Training.
    \item \autoref{supp:result} More Results.
\end{itemize}



\section{Limitations}
\label{supp:limitations}

While our framework demonstrates strong performance, several opportunities for extension remain.
Our inverse dynamics model (IDM) currently focuses on a core set of primitive actions such as \texttt{click}, \texttt{type}, \texttt{move}, and \texttt{scroll}.
It does not yet handle keyboard shortcut or hotkey based actions (\eg, \texttt{Ctrl+C} or \texttt{Ctrl+Z}), as such events are rarely captured in web based state transition data and therefore lack corresponding visual context.
Nevertheless, most hotkey operations have direct graphical interface equivalents, such as selecting menu items or clicking toolbar icons, allowing the IDM to retain near complete functional coverage of real world tasks.
Future work could incorporate application based  data collection or key event logging to expand the action space and support shortcut based interactions more efficiently.

Second, our retrieval framework retrieves demonstrations at the granularity of full tasks. While effective, this may not always align with the granularity needed by an agent during execution. Future work could explore mechanisms to automatically merge shorter tasks into longer workflows, or split lengthy tutorials into more targeted sub-trajectories. Such advances would enable finer-grained retrieval and more flexible trajectory construction, ultimately improving the adaptability of our approach.  

We view these limitations not as fundamental barriers but as natural opportunities to further enhance the scalability and generality of our framework.

\section{Dataset Details}
\label{supp:dataset}

\subsection{Applications by Category}

We selected seven categories: Productivity, Programming, Design, Screen Editing, 
Audio Production, System Utilities, and Science \& Data. These categories span a broad range of 
realistic computer use. Productivity tools (\eg, Microsoft Office, Google Workspace) 
cover everyday document and collaboration tasks, while Programming environments 
(\eg, VS Code, Jupyter) capture software development workflows. Design 
(\eg, Photoshop, Figma), Screen Editing (\eg, Premiere Pro, OBS Studio), and 
Audio Production (\eg, Audacity, FL Studio) extend to creative domains with 
specialized interfaces. System Utilities (\eg, Task Manager, Finder, Docker) 
test low-level system interaction, and Science \& Data tools (\eg, MATLAB, Tableau, SPSS) 
represent analytical and visualization tasks.  

Applications within each category were chosen for their widespread adoption, 
abundant tutorial availability on YouTube, and ability to showcase the diverse 
interaction challenges agents must master. While we focused on these applications, 
our method is not restricted to them: additional data can be generated from any 
new tutorial videos available on the web. The distribution of applications is in Table~\ref{tab_supp:apps}.

\begin{table}[t]
\centering
\caption{Applications grouped by category.}
\resizebox{1\linewidth}{!}{
\begin{tabular}{lp{10cm}}
\toprule
\textbf{Category} & \textbf{Applications} \\
\midrule
Productivity & Microsoft Office, Google Workspace, Notion, Evernote, OneNote, Trello, Asana, ClickUp, Monday.com, Slack, Microsoft Teams \\
Programming  & VS Code, PyCharm, IntelliJ IDEA, Eclipse, Android Studio, Xcode, Jupyter Notebook, Google Colab, RStudio, Sublime Text, Atom, GitHub Desktop \\
Design       & Adobe Photoshop, Adobe Illustrator, Adobe XD, Figma, Sketch, Canva, CorelDRAW, Inkscape, Gimp \\
Screen Editing & Adobe Premiere Pro, Final Cut Pro, DaVinci Resolve, Camtasia, OBS Studio, ScreenFlow, Filmora, iMovie \\
Audio Production & Audacity, Adobe Audition, FL Studio, Logic Pro X, Ableton Live, Pro Tools, Cubase, GarageBand \\
System Utilities & Windows Task Manager, PowerShell, macOS Finder, Activity Monitor, Disk Utility, Linux Terminal, Docker, VirtualBox, CCleaner, WinRAR, 7-Zip \\
Science \& Data & MATLAB, Mathematica, SPSS, SAS, Tableau, Power BI, Google Colab, Jupyter Notebook, Stata, RapidMiner \\
\bottomrule
\end{tabular}
}
\label{tab_supp:apps}
\end{table}

\section{Implementation Details}
\label{supp:implementation}

\subsection{State-Transition Data Collection}

This section provides additional details of the automated pipeline used to gather the 600K state-transition pairs $(O_t, a_t, O_{t+1})$ described in the main paper. The system consists of four components: an entry-point sampler, a browser controller, an action executor, and a transition logger. All data were collected using Playwright with a fixed viewport of $1280\times720$ pixels and 48 parallel workers.

\noindent \textbf{Entry-point sampling.}
We sample URLs from the March~2025 snapshot of the Common Crawl index~\footnote{\url{https://commoncrawl.org/blog/march-2025-crawl-archive-now-available}} (CC-MAIN-2025-13). We retain only HTML documents, filter out pages smaller than 5~KB, and discard URLs that trigger login walls or CAPTCHA challenges. To reduce domain imbalance, we cap the number of visits to 30 pages per domain. Each browsing session begins from a randomly chosen entry point, ensuring high visitation diversity across the crawl.

\noindent \textbf{Action sampling policy.}
At each timestep, the pipeline samples an action from our six primitives: \texttt{click}, \texttt{release}, \texttt{scroll}, \texttt{type}, \texttt{wait}, and \texttt{move}. We empirically bias the sampling distribution toward common web interactions: click ($\sim$45\%), scroll ($\sim$20\%), move ($\sim$15\%), type ($\sim$8\%), wait ($\sim$2\%), and release ($\sim$10\%). Coordinates for \texttt{click} and \texttt{move} actions are drawn uniformly over the viewport and jittered around visible UI elements obtained from Playwright’s accessibility tree. Scroll magnitudes follow a fixed value (500\,px). Text for \texttt{type} actions is sampled from a curated list of $\sim$5K frequently used English words~\footnote{\url{https://www.vocabulary.com/lists/154147}}.

\noindent \textbf{Transition recording.}
For each action, we record the pre- and post-action screenshots $(O_t, O_{t+1})$, the action primitive, all associated parameters, the current page URL, and a timestamp. Transitions where $O_t$ and $O_{t+1}$ differ by fewer than 50 nonzero pixels (excluding \texttt{wait}) are removed to avoid degenerate interactions. Each action is executed with a 5\,s timeout; failed executions or delayed page loads are logged and the transition is discarded.

\noindent \textbf{Error handling and session resets.}
The system automatically detects HTTP~4xx/5xx responses, navigation loops, frozen page states, and browser crashes. Upon such events, the session is terminated and restarted with a fresh entry point. To promote diversity, each session is limited to 15 actions before being reset, preventing oversampling from any single page.

\noindent \textbf{Scale and runtime.}
Data collection was run on CPU-only machines (32 cores, 64\,GB RAM). Across 48 parallel workers, the pipeline collected approximately 600K high-quality transitions over 3.5 days. After filtering, the resulting dataset covers a wide range of layouts, domains, and dynamic interaction patterns encountered in real web environments.

\subsection{Video Retrieval}

To build a large-scale dataset of application demonstrations, we require a method 
to identify relevant tutorial videos from the web. YouTube is a natural source 
since it contains abundant tutorials across productivity, programming, design, 
and other domains. However, naively searching by task description may yield 
irrelevant or entertainment-focused videos. To address this, we designed a 
dedicated prompt for generating targeted search queries.  

The prompt (shown below) instructs a language model to act as an expert in 
YouTube search, taking as input a task description and a list of related 
applications. It outputs a short and effective query that emphasizes tutorials, 
how-to videos, and instructional content. By constraining queries to be concise 
and domain-specific, this approach improves retrieval precision and reduces 
noise from unrelated videos.

\begin{tcolorbox}[colframe=blue!50!black, colback=blue!10!white, title=Prompt for Video Retreival Query Generation, breakable]

\small
You are an expert at creating YouTube search queries. Given a task instruction and related applications, 
create a concise, effective search query that will find relevant tutorial videos.

\smallskip

Task: \{instruction\}

Related Applications: \{related\_apps\}

\smallskip

Create a search query that would find helpful tutorial videos for this task. 
Focus on tutorial, how-to, or instructional content. Keep it concise (under 10 words).

Search query:

\end{tcolorbox}

\subsection{Video Filtering}

After retrieving candidate tutorials, many videos still contain irrelevant or 
low-quality content such as talking-head introductions, presentation slides, or 
animated transitions. To ensure that our dataset is composed of high-quality 
screen recordings that clearly demonstrate application use, we apply a filtering 
step.  

We design a prompt that instructs a language model to act as a visual classifier. 
Given a single frame from a video, the model assigns both a categorical label 
(\eg, clean screencast, zoomed screencast, talking head) and a quality score 
between 0.0 and 1.0. We retain only those videos where the average frame score 
exceeds 0.8, which empirically yields a reliable set of clean tutorial 
screencasts. This threshold balances recall and precision: it removes noisy or 
non-screencast content while retaining a broad coverage of genuine tutorials.  

\begin{tcolorbox}[colframe=blue!50!black, colback=blue!10!white, title=Prompt for Video Filtering, breakable]
\small

You are a visual classifier helping to filter video tutorial frames for clean screencast content.

Your task is to classify an input image (a single frame from a video) and provide a quality score.

\medskip

Classify the image into one of these categories:

1. Clean Screencast: Full desktop screen showing software interface, application window, code editor, browser, or terminal. Clear, unzoomed view of the entire screen or application window.

\smallskip

2. Zoomed Screencast: Screenshot that has been zoomed in or cropped, showing only part of the screen or interface elements.

\smallskip

3. Animated/Transition: Frames with animations, transitions, intro/outro effects, or visual effects that are not static screencast content.

\smallskip

4. Talking Head: Person's face or upper body from webcam, typically in corner or overlay.

\smallskip

5. Slide/Presentation: Static presentation slide, diagram, or text-heavy content.

\smallskip

6. Other: Content that doesn't fit the above categories.

\medskip

For each classification, also provide a quality score from 0.0 to 1.0:
- 1.0: Perfect clean screencast
- 0.8-0.9: Good screencast with minor issues
- 0.6-0.7: Acceptable screencast
- 0.4-0.5: Poor quality or partially zoomed
- 0.0-0.3: Very poor or not screencast

\medskip

Return your response in this format:
Category: [category name]
Quality: [score]
Reason: [brief explanation]

\end{tcolorbox}

\subsection{In-context Learning}

To evaluate \ourmethod{} under in-context learning (ICL), we augment the default prompts used by state-of-the-art OSWorld agents with a small number of structured demonstration trajectories. The OSWorld repository provides model-specific system prompts in OS-World Github~\footnote{\url{https://github.com/xlang-ai/OSWorld/blob/main/mm_agents/prompts.py}}, which define the action format, reasoning style, and interface conventions. We prepend our exemplars to these prompts without modifying any downstream agent logic.

For a given evaluation task, we retrieve three relevant trajectories from our video corpus and convert each into a sequence of \textit{(observation, thought, action)} triples. These triples are formatted using the exact conventions defined by OSWorld agents, including the same observation style, reasoning structure, and action specification. We prepend the resulting exemplars to the agent prompt by placing them inside a dedicated instruction block that appears immediately \emph{before} the OSWorld rules and reasoning instructions. This placement ensures that the agent first learns the desired behavior pattern from the exemplars, and then applies the original constraints that follow.

\begin{tcolorbox}[colframe=blue!50!black, colback=blue!10!white, title=ICL Example Prompt Block, breakable]

\textbf{Example Trajectories}

Below are some example task trajectories showing how to observe, think, and act. 
Study these examples carefully to understand:
\begin{itemize}
    \item how to provide detailed observations of the screen state,
    \item how to structure your reasoning process step by step,
    \item how to format action code with clear comments,
    \item how to choose appropriate coordinates based on visual elements,
    \item when to use special codes like \texttt{WAIT}, \texttt{DONE}, or \texttt{FAIL},
    \item how to recover from errors by adjusting your approach.
\end{itemize}

\smallskip
These examples demonstrate similar tasks to what you'll be performing. Use them as a reference for the level of detail expected in your observations and the reasoning process for choosing actions. Pay attention to how coordinates are selected based on element positions and how actions are adapted when previous attempts don't succeed.

\medskip
Example \texttt{\{example\_num\}}: \texttt{\{task\_description\}}

Step \texttt{\{step\_num\}}:\\
Screenshot: \texttt{[Image showing the current state]}

Thought: \texttt{\{thought\_text\}}

Action:
\begin{verbatim}
{action_code}
\end{verbatim}

\end{tcolorbox}

\noindent \textbf{Example of Full ICL Prompt with O3 Agent.}
In ~\cref{supp:fig:o3_prompt} we show how a set of exemplar trajectories is prepended to the O3 agent instructions. The exemplars appear after the initial instruction, followed immediately by the rule and reasoning sections used by the OSWorld agent.

\subsection{Models}
For in-context learning evaluations we query API-based models using their latest public versions: Google Gemini~2.5 Flash (\texttt{gemini-2.5-flash}), OpenAI o3 (\texttt{o3-2025-04-16}), and Anthropic Claude~4 Sonnet (\texttt{claude-4-sonnet-20250514}). We use deterministic decoding with temperature set to 0.0.

For IDM training, we use the AdamW optimizer with a learning rate of $3\mathrm{e}{-4}$, batch size 256, and cosine learning rate decay. Training is run for 15 epochs on 8$\times$A100 GPUs (80GB) with gradient clipping at 1.0 and mixed-precision (bfloat16). For supervised fine-tuning of CUAs, we follow the official training recipes from UI-TARS-1.5 and Qwen~2.5-VL, adapting batch size to fit the same hardware setup.

\subsection{Action Space and Primitive Composition}

The inverse dynamics model (IDM) predicts atomic GUI interaction primitives from observed state transitions. The primitive action space consists of \textit{Click}, \textit{Release}, \textit{Move}, \textit{Scroll}, \textit{Type}, and \textit{Wait}. These primitives serve as a low-level interaction vocabulary and are not executed directly in the environment.

Instead, predicted primitive sequences are converted into executable environment actions through a deterministic composition layer implemented using \texttt{pyautogui}. This layer maps temporally ordered primitive sequences to environment-compatible commands. For instance, the sequence \texttt{Click$\rightarrow$Release} corresponds to a standard mouse click, while \texttt{Click$\rightarrow$Move$\rightarrow$Release} corresponds to drag-based interactions such as \texttt{moveto} or \texttt{dragto}. The composition stage enforces a consistent interface between the model output space and the environment action space.

Primitive distinctions are introduced only when they induce observable changes in the visual state. When multiple interaction sequences are visually indistinguishable under the observation stream, they are resolved to the minimal primitive sequence consistent with the observed transition. This constraint avoids introducing latent action distinctions that cannot be inferred from visual evidence.

To ensure compatibility with downstream execution, the IDM primitive space is aligned with a subset of the agent and environment action spaces through the deterministic mapping described above. This alignment constrains the prediction space to executable interactions and mitigates degenerate solutions during training by preventing the model from collapsing onto non-executable or redundant action representations.

\section{Inverse Dynamics Model Architecture}
\label{supp:idm_arch}
\subsection{Vision Encoder}
\label{sec:vision_encoder}

\noindent \textbf{Model Selection.} 
We employ SigLIP-2 Base (patch size 16) with the NaFlex variant as our vision encoder~\cite{tschannen2025siglip}. This architecture was specifically chosen for its superior performance on OCR and document-understanding benchmarks, which closely aligns with the requirements of computer interface understanding where precise text recognition and spatial localization are critical.

\noindent \textbf{Input Processing.} 
Each observation $O_t$ is a 1000$\times$1000 RGB screenshot. We utilize the NaFlex preprocessing strategy with \texttt{max\_num\_patches=1024}, which preserves the square aspect ratio while resizing the input to approximately 704$\times$704 pixels before patch extraction. This configuration yields exactly 1024 visual tokens (a 32$\times$32 grid of 16$\times$16 patches), providing sufficient spatial resolution for identifying small UI elements such as buttons, icons, and text labels while maintaining computational efficiency.

\noindent \textbf{Feature Extraction.} 
For the pair $(O_t, O_{t+1})$, we process each observation independently through the frozen SigLIP-2 encoder to obtain visual embeddings $\mathbf{v}_t, \mathbf{v}_{t+1} \in \mathbb{R}^{1024 \times 768}$. These embeddings preserve spatial information critical for coordinate prediction.

\noindent \textbf{Rationale for NaFlex.} 
Compared to the fixed-resolution SigLIP-2 variant at 512$\times$512 (which would downsample our 1000$\times$1000 input by 48\%), NaFlex provides:
(1) Higher effective resolution ($\sim$704$\times$704) for better preservation of fine-grained UI details,
(2) Superior performance on text-heavy and structured visual content~\cite{tschannen2025siglip}, and
(3) Minimal aspect ratio distortion for square inputs.
While fixed-resolution variants benefit from additional distillation stages, preliminary experiments showed NaFlex's advantages in OCR-like capabilities outweigh this trade-off for UI understanding.

\subsection{Transformer Backbone}
\label{sec:transformer_backbone}

\noindent \textbf{Temporal Feature Fusion.} 
The patch-level features from $(O_t, O_{t+1})$ are concatenated along the sequence dimension, yielding a combined representation of shape $\mathbb{R}^{2048 \times 768}$. We prepend a learnable \texttt{[CLS]} token to this sequence. To distinguish temporal information, we add learnable temporal embeddings $\mathbf{e}_t, \mathbf{e}_{t+1} \in \mathbb{R}^{768}$ to the respective halves of the sequence.

\noindent \textbf{Architecture.} 
Our backbone consists of 4 transformer layers with the following specifications:
\begin{itemize}
    \item \textbf{Hidden dimension}: 768 (matching SigLIP-2 output)
    \item \textbf{Attention heads}: 12
    \item \textbf{MLP hidden dimension}: 3072 (4$\times$ expansion)
    \item \textbf{Activation function}: GELU
    \item \textbf{Dropout}: 0.1 applied to attention weights and MLP outputs
    \item \textbf{Layer normalization}: Pre-norm configuration with $\epsilon=10^{-6}$
    \item \textbf{Attention mechanism}: Standard scaled dot-product attention with causal masking disabled (full bidirectional attention)
\end{itemize}

The final \texttt{[CLS]} token representation $\mathbf{h}_{\text{cls}} \in \mathbb{R}^{768}$ serves as the global action embedding, while the patch-level outputs $\mathbf{h}_{\text{patch}} \in \mathbb{R}^{2048 \times 768}$ are used for coordinate prediction.

\subsection{Prediction Heads}
\label{sec:prediction_heads}

\subsubsection{Action Classification Head}

A simple two-layer MLP predicts the action type:
\begin{equation}
\mathbf{p}_{\text{action}} = \text{softmax}(\mathbf{W}_2 \cdot \text{ReLU}(\mathbf{W}_1 \mathbf{h}_{\text{cls}} + \mathbf{b}_1) + \mathbf{b}_2)
\end{equation}
where $\mathbf{W}_1 \in \mathbb{R}^{512 \times 768}$, $\mathbf{W}_2 \in \mathbb{R}^{6 \times 512}$, and $\mathbf{p}_{\text{action}} \in \mathbb{R}^6$ represents the probability distribution over the six action primitives: \texttt{click}, \texttt{release}, \texttt{scroll}, \texttt{type}, \texttt{wait}, and \texttt{move}.

\subsubsection{Coordinate Prediction Head}

\noindent \textbf{Architecture.} 
For location-based actions, we predict normalized coordinates discretized into 1000 bins for each axis. We employ separate prediction heads for $x$ and $y$ coordinates:

\noindent For the $x$-coordinate:
\begin{align}
\mathbf{h}_x &= \text{AvgPool}(\mathbf{h}_{\text{patch}}) \quad \text{(over spatial dimension)} \\
\mathbf{p}_x &= \text{softmax}(\mathbf{W}_x \cdot \text{ReLU}(\text{LayerNorm}(\mathbf{h}_x)))
\end{align}

where $\mathbf{W}_x \in \mathbb{R}^{1000 \times 768}$ and $\mathbf{p}_x \in \mathbb{R}^{1000}$. The $y$-coordinate head is parameterized identically with separate weights $\mathbf{W}_y$.

\noindent \textbf{Discretization Strategy.} 
Screen coordinates are normalized to $[0, 1]$ and discretized into 1000 uniform bins. This formulation offers several advantages:
(1) Transforms regression into classification, enabling the use of cross-entropy loss which empirically provides stronger gradients than regression losses (\emph{e.g.}, L1, L2),
(2) Naturally handles multi-modal distributions when multiple valid click locations exist,
(3) Provides calibrated uncertainty estimates via the output probability distribution.

\noindent \textbf{Resolution Alignment.} 
The 1000-bin discretization aligns naturally with our 1000$\times$1000 input resolution, where each bin corresponds to approximately a 1-pixel region. This granularity is sufficient for UI interaction, as most clickable targets span multiple pixels.

\subsubsection{Language Head}

\noindent \textbf{Architecture.} 
For text-entry actions (\texttt{type}), we employ a lightweight autoregressive decoder based on GPT-2~\cite{gpt2}.


\noindent \textbf{Visual Conditioning.} 
The decoder attends to the visual features via cross-attention. Specifically, at each decoder layer $\ell$, we insert a cross-attention sublayer:
\begin{equation}
\text{CrossAttn}^{(\ell)}(Q, K, V) = \text{softmax}\left(\frac{Q K^T}{\sqrt{d_k}}\right) V
\end{equation}
where $Q$ is derived from the decoder's hidden states, and $K, V$ are projections of the concatenated visual features $[\mathbf{v}_t; \mathbf{v}_{t+1}]$ from the SigLIP-2 encoder. This design enables the model to ground its text generation in the observed visual transition.

\noindent \textbf{Training Strategy.} 
During training, we use teacher forcing with cross-entropy loss on the ground-truth text sequence.

\section{Inverse Dynamics Model Training}
\label{supp:idm_training}
This part provides the full definition of the training losses used for
the inverse dynamics model (IDM) in \ourmethod{}.  
The IDM predicts three types of supervision depending on the action:
an action primitive, optional pixel coordinates, and optional text tokens.
For completeness, we describe the outputs, losses, and training objective
in detail below.

\subsection{Model Outputs}

Given two consecutive observations $(O_t, O_{t+1})$, where $O_t$ and
$O_{t+1}$ are screen images before and after executing action $a_t$, the
IDM produces the following outputs:

\begin{itemize}
    \item \textbf{Action primitive}:  
    A categorical distribution $p_\theta(a_t)$ over the set of supported
    primitives (e.g., \texttt{click}, \texttt{scroll}, \texttt{type},
    \texttt{wait}, \texttt{move}).

    \item \textbf{Coordinate predictions}:  
    For actions requiring a spatial position, the model predicts
    distributions $p_\theta(x_t)$ and $p_\theta(y_t)$ over discretized
    pixel bins $\{0,\ldots,999\}$.

    \item \textbf{Typed text}:  
    For text-entry actions, the IDM predicts the sequence of typed tokens
    $w_{1:L}$ autoregressively through
    $p_\theta(w_\ell \mid w_{<\ell}, O_t, O_{t+1})$.
\end{itemize}

\begin{table*}[t]
\centering
\caption{Detailed OSWorld category-wise task successes. \ourmethod{} provides the strongest improvements in domains with abundant specialized tutorials (\eg, Chrome, Gimp, VLC), while gains are smaller in domains requiring heavy text entry, rare actions, or fine-grained control.}
\resizebox{\textwidth}{!}{
\begin{tabular}{c c l c c c c c c c c c c c}
\toprule
Setting & Category & Model & chrome & gimp & lo\_calc & lo\_impress & lo\_writer & multi\_apps & os & thunderbird & vlc & vs\_code & Total \\
\midrule
\multirow{8}{*}{ICL} 
  & \multirow{6}{*}{\shortstack{General \\ Models}} 
    & Gemini 2.5 Flash & 8 & 8 & 4 & 3 & 5 & 9 & 10 & 6 & 5 & 12 & 70 \\
  &   & + \textbf{\ourmethod} & \textbf{10 (+3)} & \textbf{10 (+2)} & 4  & 5  & 5 & 9 & 10 & \textbf{8 (+2)} & 8 (+3) & 12 & \textbf{81 (+11)} \\
  &   & o3 & 6 & 10 & 5 & 5 & 7 & 15 & 15 & 4 & 7 & 9 & 83 \\
  &   & \textbf{+ \ourmethod} & \textbf{9 (+3)} & \textbf{13 (+2)} & \textbf{7 (+1)} & 7 & 7 & \textbf{18 (+1)} & 15 & 4 & \textbf{9 (+2)} & 9 & \textbf{98 (+9)} \\
  &   & Claude 4 Sonnet & 25 & 13 & 15 & 22 & 14 & 27 & 11 & 11 & 7 & 14 & 159 \\
  &   & \textbf{+ \ourmethod} & \textbf{27 (+2)} & \textbf{15 (+2)} & 15 & 22 & 14 & 27 & 11 & 11 & \textbf{9 (+2)} & 14 & \textbf{169 (+6)} \\
\cmidrule{2-14}
  & \multirow{2}{*}{\shortstack{Agentic \\ Framework}} 
    & Jedi & 26 & 21 & 19 & 21 & 15 & 32 & 13 & 12 & 10 & 13 & 182 \\
    &   & \textbf{+ \ourmethod} & \textbf{29 (+3)} & \textbf{23 (+2)} & 19 & \textbf{23 (+2)} & 15 & 32 & 13 & 12 & \textbf{12 (+2)} & 13 & \textbf{191 (+9)} \\
\midrule
\multirow{4}{*}{SFT} 
  & \multirow{4}{*}{\shortstack{Open-Weight \\ Models}} 
    & UI-TARS-1.5-7B & 11 & 15 & 6 & 14 & 9 & 5 & 8 & 4 & 6 & 15 & 93 \\
    &      & \textbf{+ \ourmethod} & \textbf{13 (+2)} & \textbf{17 (+2)} & \textbf{8 (+2)} & \textbf{16 (+2)} & 9 & \textbf{7 (+2)} & 8 & \textbf{4 (+2)}  & \textbf{7 (+2)} & 15 & \textbf{104 (+14)} \\
  &   & Qwen 2.5-VL 7B & 4 & 1 & 0 & 0 & 2 & 0 & 0 & 2 & 2 & 0 & 7 \\
  &   & + \textbf{\ourmethod} & \textbf{12 (+8)} & \textbf{10 (+9)} & \textbf{3 (+3)} & \textbf{1 (+1)} & 2 & \textbf{1 (+1)} & \textbf{5 (+5)} & \textbf{4 (+2)} & \textbf{6 (+4)} & \textbf{4 (+4)} & \textbf{48 (+41)} \\
\bottomrule
\end{tabular}}
\label{tab:domain_breakdown}
\end{table*}

Each action is routed only to the heads that apply to its supervision.

\subsection{Overall Objective}

The entire model is trained end-to-end using a multi-task objective that
adds the losses from the applicable prediction heads:
\begin{equation}
\mathcal{L}
= \lambda_{\mathrm{cls}} \mathcal{L}_{\mathrm{cls}}
+ \lambda_{\mathrm{coord}} \mathcal{L}_{\mathrm{coord}}
+ \lambda_{\mathrm{lang}} \mathcal{L}_{\mathrm{lang}},
\end{equation}
where $\lambda_{\mathrm{cls}}$, $\lambda_{\mathrm{coord}}$, and
$\lambda_{\mathrm{lang}}$ are scalar weights (set to $1$ by default).
Loss terms are included only for actions with the corresponding
supervision signal.

\subsubsection{Classification Loss}

Every action includes a primitive label
$a_t^{\mathrm{cls}}$, which is supervised using a standard cross-entropy
loss:
\begin{equation}
\mathcal{L}_{\mathrm{cls}}
= -\log p_\theta(a_t^{\mathrm{cls}} \mid O_t, O_{t+1}).
\end{equation}
This term teaches the IDM to infer the underlying categorical action from
the state transition.

\subsubsection{Coordinate Loss}

For location-based actions (such as \texttt{click}, \texttt{move}, or
\texttt{type}), the IDM predicts normalized pixel coordinates
$(x_t, y_t)$, each discretized into $1000$ bins.
We define the coordinate loss as the sum of independent cross-entropy
terms for the two axes:
\begin{equation}
\mathcal{L}_{\mathrm{coord}}
= -\log p_\theta(x_t) - \log p_\theta(y_t).
\end{equation}
This term is omitted for actions that do not require coordinates.

\subsubsection{Language Modeling Loss}

For text-entry actions, the IDM predicts the token sequence $w_{1:L}$
using a left-to-right autoregressive decoder.  
The text loss is the usual negative log-likelihood over the sequence:
\begin{equation}
\mathcal{L}_{\mathrm{lang}}
= -\sum_{\ell=1}^{L}
\log p_\theta(w_\ell \mid w_{<\ell}, O_t, O_{t+1}),
\end{equation}
where $L$ is the length of the typed sequence.
This loss is skipped entirely for non-text actions.

\bigskip

Together, these losses encourage the IDM to reconstruct action primitives,
spatial arguments, and typed text directly from observed UI transitions,
providing a unified formulation for all interaction types.

\section{More Results}
\label{supp:result}

\subsection{What is the effect of data scale for supervised fine-tuning?}

\begin{table}[h]
\centering
\caption{Data scaling results on OSWorld with Qwen 2.5-VL. Performance improves as training data increases from 10k to 25k and the full dataset.}
\begin{tabular}{l c c c c}
\toprule
Model & Base & 10k & 25k & Full \\
\midrule
Qwen 2.5-VL & 1.9 & 3.3 & 4.9 & 13.0 \\
\bottomrule
\end{tabular}
\label{tab_supp:qwen_data_scaling}
\end{table}

We study how scaling the number of training trajectories affects the performance of Qwen 2.5-VL on OSWorld. As shown in Table~\ref{tab_supp:qwen_data_scaling}, success rates increase from 1.9\% with the base model to 3.3\% with 10k trajectories, 4.9\% with 25k trajectories, and 13.0\% with the full dataset. The improvement is closer to exponential than linear, suggesting that a minimum critical mass of data is required before substantial gains emerge. 

We hypothesize that this behavior arises because Qwen must learn both \emph{grounding} and \emph{planning} from the video-derived trajectories. With limited data, the model struggles to acquire either capability robustly, leading to only small improvements. Once enough trajectories are available, however, Qwen begins to effectively integrate grounding of UI states with coherent planning patterns, producing sharper gains. This indicates that further scaling of high-quality trajectories could unlock even larger benefits.

\subsection{Which application domains benefit most from our data?}

To better understand the strengths and limitations of our approach, we break down results by application domain on OSWorld. Table~\ref{tab:domain_breakdown} reports task successes for general-purpose models (o3, Claude~4 Sonnet), the Jedi agentic framework, and the open-source model UI-TARS-7B, both with and without \ourmethod{} exemplars or training data.  

The largest improvements are observed in \texttt{chrome}, \texttt{gimp}, and \texttt{vlc}. These domains benefit strongly from specialized procedural knowledge that is well covered by online tutorials, such as configuring browser settings, editing images, or adjusting media player preferences. The presence of abundant, step-by-step demonstrations in these categories enables our pipeline to extract high-quality trajectories that transfer effectively to downstream agents.  

By contrast, the gains are smaller in domains such as \texttt{vscode} and \texttt{os}, which often require extensive text entry or code manipulation---capabilities that are less easily captured by our current action set. Improvements are also limited in \texttt{thunderbird} and LibreOffice applications (\texttt{lo.calc}, \texttt{lo.writer}, \texttt{lo.impress}), where high-quality tutorials are scarce and tasks sometimes involve fine-grained interactions such as dragging objects or manipulating small interface elements.

Overall, this breakdown highlights a key property of our approach: it yields the largest benefits in domains where web tutorials are both plentiful and aligned with the action space of the agent, while leaving room for future extensions in text-heavy or fine-grained interaction domains.

\clearpage
\begin{figure*}[t!]
\begin{tcolorbox}[width=\linewidth, enhanced jigsaw, colframe=blue!50!black, colback=blue!10!white, title=Combined ICL Prompt for o3]

You are an agent which follow my instruction and perform desktop computer tasks as instructed.
You have good knowledge of computer and good internet connection and assume your code will run on a computer for controlling the mouse and keyboard.
You are on Ubuntu operating system and the resolution of the screen is 1920x1080.
For each step, you will receive a screenshot of the current computer screen and must predict the corresponding action.

\medskip

\textbf{Example Trajectories}

Below are some example task trajectories showing how to observe, think, and act. 
Study these examples carefully to understand:
\begin{itemize}
    \item how to provide detailed observations of the screen state,
    \item how to structure your reasoning process step by step,
    \item how to format action code with clear comments,
    \item how to choose appropriate coordinates based on visual elements,
    \item when to use special codes like \texttt{WAIT}, \texttt{DONE}, or \texttt{FAIL},
    \item how to recover from errors by adjusting your approach.
\end{itemize}

\smallskip
These examples demonstrate similar tasks to what you will be performing. Use them as a reference for the level of detail expected in your observations and the reasoning process for choosing actions.

\medskip
\textbf{Example 1: Open a settings panel}

Step 1:\\
Screenshot: \texttt{[Image showing desktop with a gear-shaped Settings icon in the left sidebar]}

Thought: \texttt{The Settings icon is visible in the sidebar. Clicking it should open the system settings window.}

Action:
\begin{verbatim}
# Click the gear-shaped "Settings" icon in the left sidebar
pyautogui.click(x=120, y=340, duration=1)
\end{verbatim}


\medskip
\textbf{The following rules are important:}
\begin{itemize}
    \item If previous actions did not achieve the expected result, do not repeat them. Adjust your action or coordinates based on the new screenshot.
    \item Do not predict multiple clicks at once. Base each action only on what is currently visible.
    \item You cannot complete the task by outputting text. You must interact through mouse and keyboard. Return \texttt{Fail} when the task cannot be done.
\end{itemize}

\medskip
\textbf{Observation.}
Provide a detailed observation of the current computer state based on the full screenshot in an ``Observation:'' section. Include any potentially relevant information such as pop-ups, notifications, or error messages.  
You \emph{must} output the observation before the thought.

\medskip
\textbf{Thought.}
Think step by step and describe the full reasoning process behind your next action. Your thought should include:
\begin{itemize}
    \item a progress assessment and reflection on previous attempts,
    \item identification of errors or mismatches,
    \item enumeration of possible next actions based on visible state,
    \item analysis and selection of the best next action.
\end{itemize}
You \emph{must} output the thought before the action code.

\medskip

\textbf{Action Format.}
Use \texttt{pyautogui} to perform actions, but do \textbf{not} use \texttt{pyautogui.locateCenterOnScreen} or \texttt{pyautogui.screenshot()}.  
Return exactly one line of Python code, preceded by a clear instruction comment. 
Return the code inside a code block:
\begin{verbatim}
```python
# instruction
pyautogui.click(x=..., y=..., duration=1)
\end{verbatim}

\end{tcolorbox}
\label{supp:fig:o3_prompt}
\end{figure*}

\begin{figure*}[h]
\begin{tcolorbox}[width=\linewidth, enhanced jigsaw, colframe=blue!50!black, colback=blue!10!white, title=Combined ICL Prompt for o3 (continued)]

\medskip

\textbf{Special tokens.}
\begin{itemize}
    \item \texttt{WAIT}: when a pause is needed,
    \item \texttt{DONE}: when the task is completed,
    \item \texttt{FAIL}: when the task cannot be completed.
\end{itemize}
\medskip
You have a maximum of 100 steps and the current step is \{current\_step\} out of \{max\_steps\}.
My computer's password is '\{CLIENT\_PASSWORD\}'. Use it when necessary for sudo operations.
First, briefly reflect on the current screenshot and prior actions, then return only the code or special token.
\end{tcolorbox}
\end{figure*}

\begin{table*}[h]
\caption{\textbf{\ourmethod{} labeling example}}
\centering
\begin{tabular}{c}
\toprule
Task: Resize image in GIMP \\
\midrule
\textbf{Observation 1 (GIMP)} \\
\midrule
\includegraphics[width=0.5\textwidth]{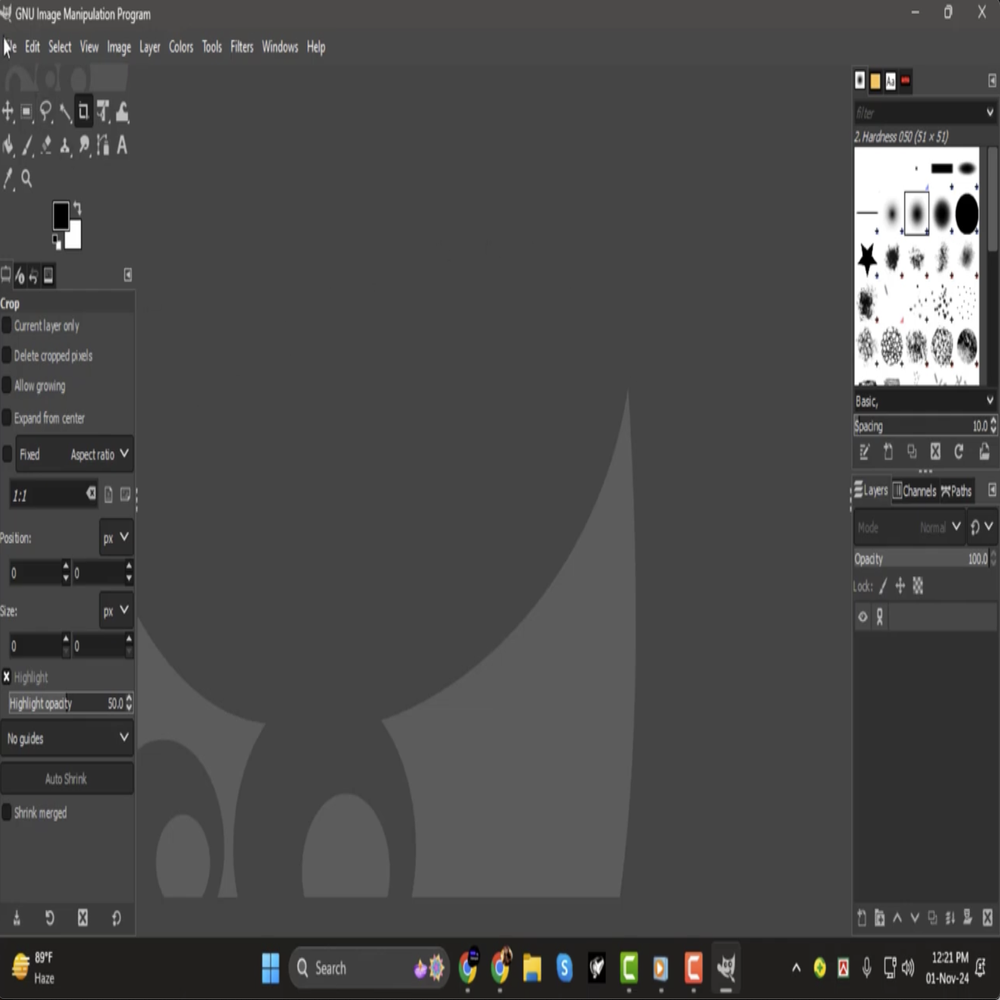} \\
\midrule
\textbf{Action 1} \\
\midrule
click (8, 45) \\
\midrule
\textbf{Observation 2 (GIMP)} \\
\midrule
\includegraphics[width=0.5\textwidth]{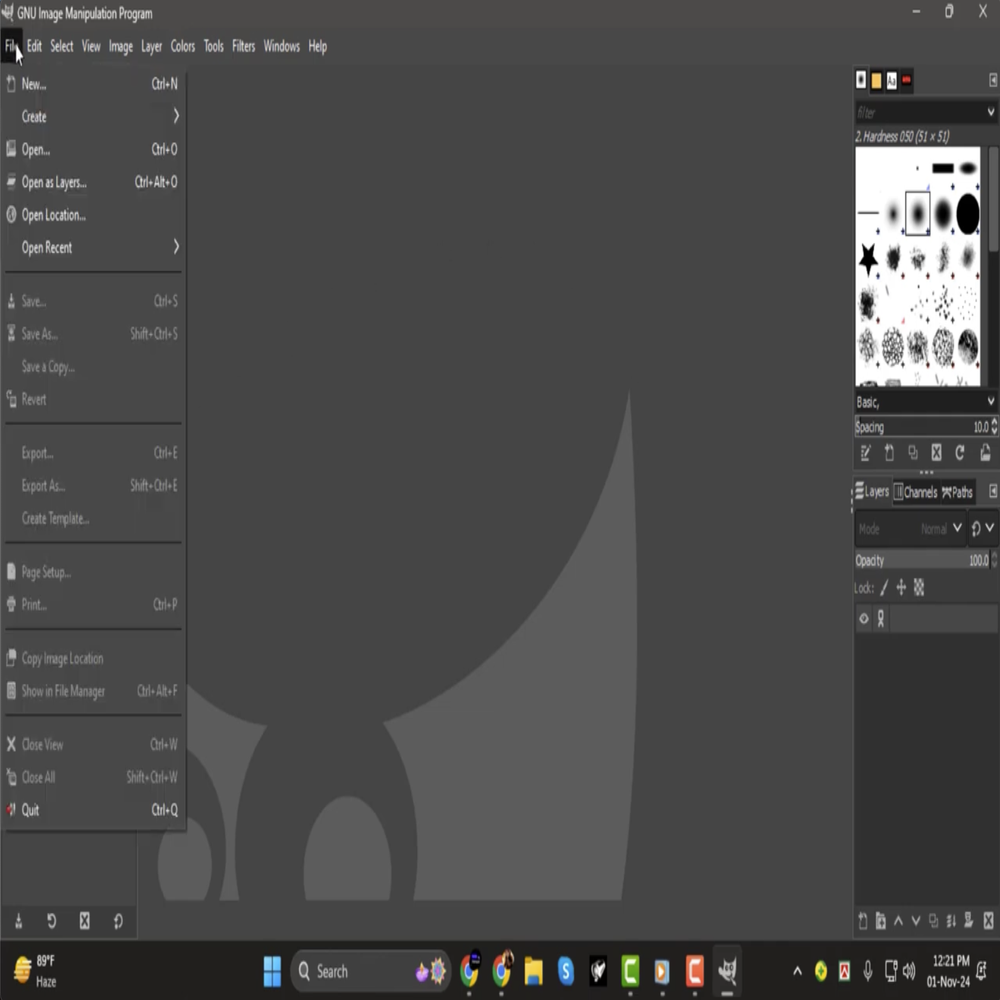} \\
\textbf{Action 2} \\
\midrule
move (34, 147) \\
\midrule
\bottomrule
\end{tabular}

\label{tab:idm_traj1_0}
\end{table*}

\begin{table*}[h]
\caption{\textbf{\ourmethod{} labeling example}}
\centering
\begin{tabular}{c}
\toprule
\textbf{Observation 3 (GIMP)} \\
\midrule
\includegraphics[width=0.5\textwidth]{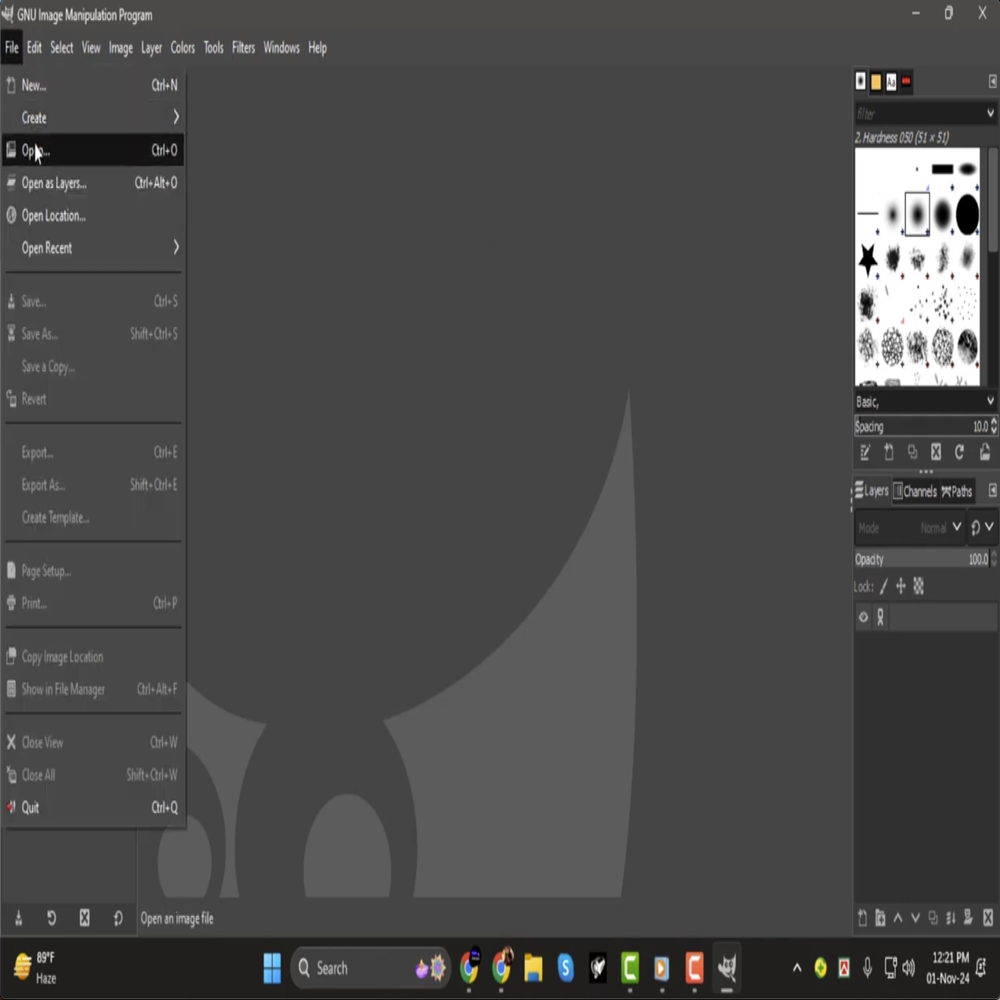} \\
\midrule
\textbf{Action 3} \\
\midrule
click (34, 147) \\
\midrule
\textbf{Observation 4 (GIMP)} \\
\midrule
\includegraphics[width=0.5\textwidth]{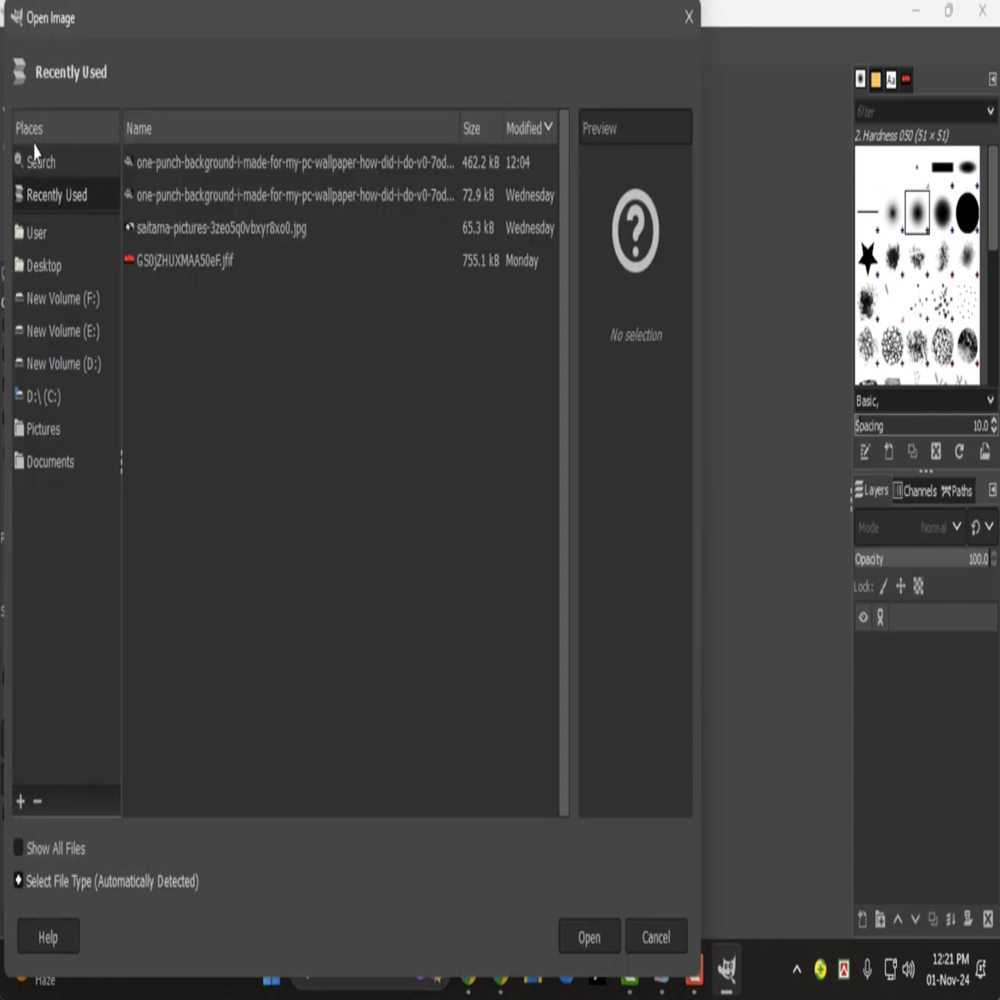} \\
\textbf{Action 4} \\
\midrule
click (75, 426) \\
\midrule
\bottomrule
\end{tabular}

\label{tab:idm_traj1_1}
\end{table*}

\begin{table*}[h]
\caption{\textbf{\ourmethod{} labeling example}}
\centering
\begin{tabular}{c}
\toprule
\textbf{Observation 5 (GIMP)} \\
\midrule
\includegraphics[width=0.5\textwidth]{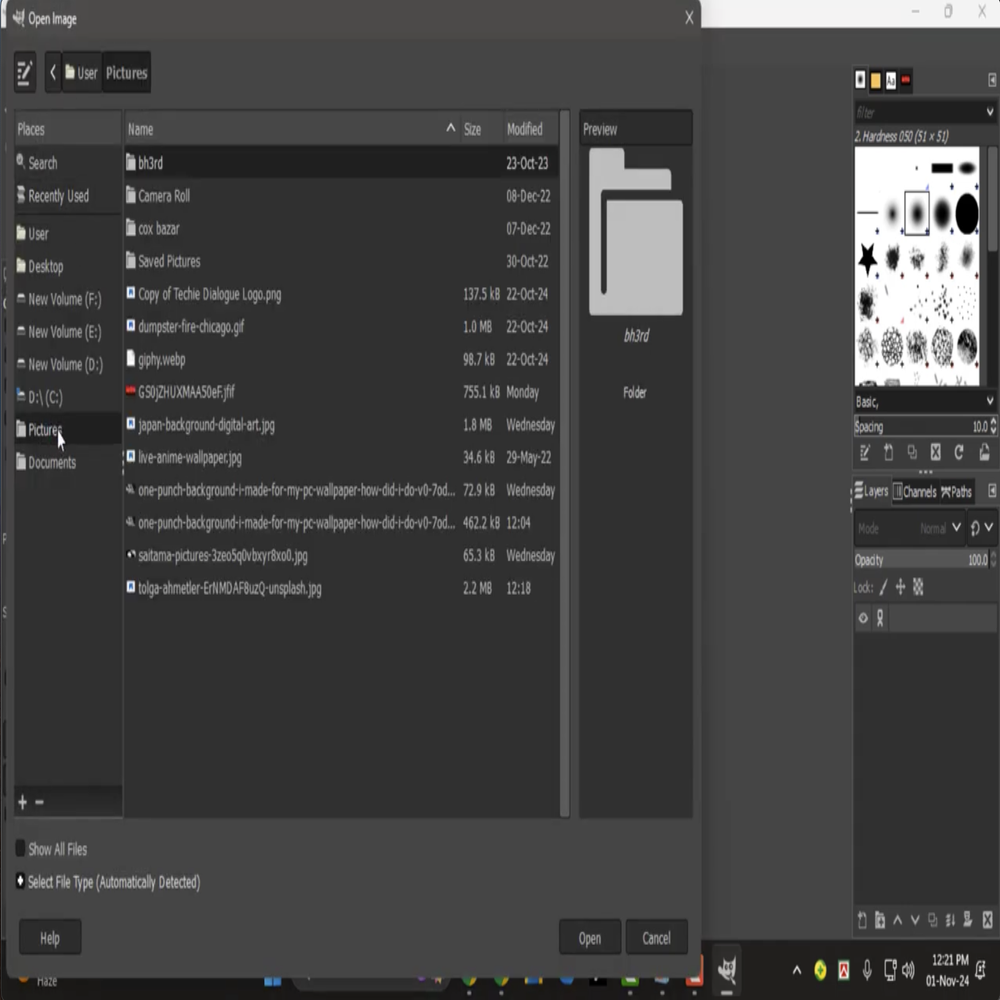} \\
\midrule
\textbf{Action 5} \\
\midrule
click (200, 594) \\
\midrule
\textbf{Observation 6 (GIMP)} \\
\midrule
\includegraphics[width=0.5\textwidth]{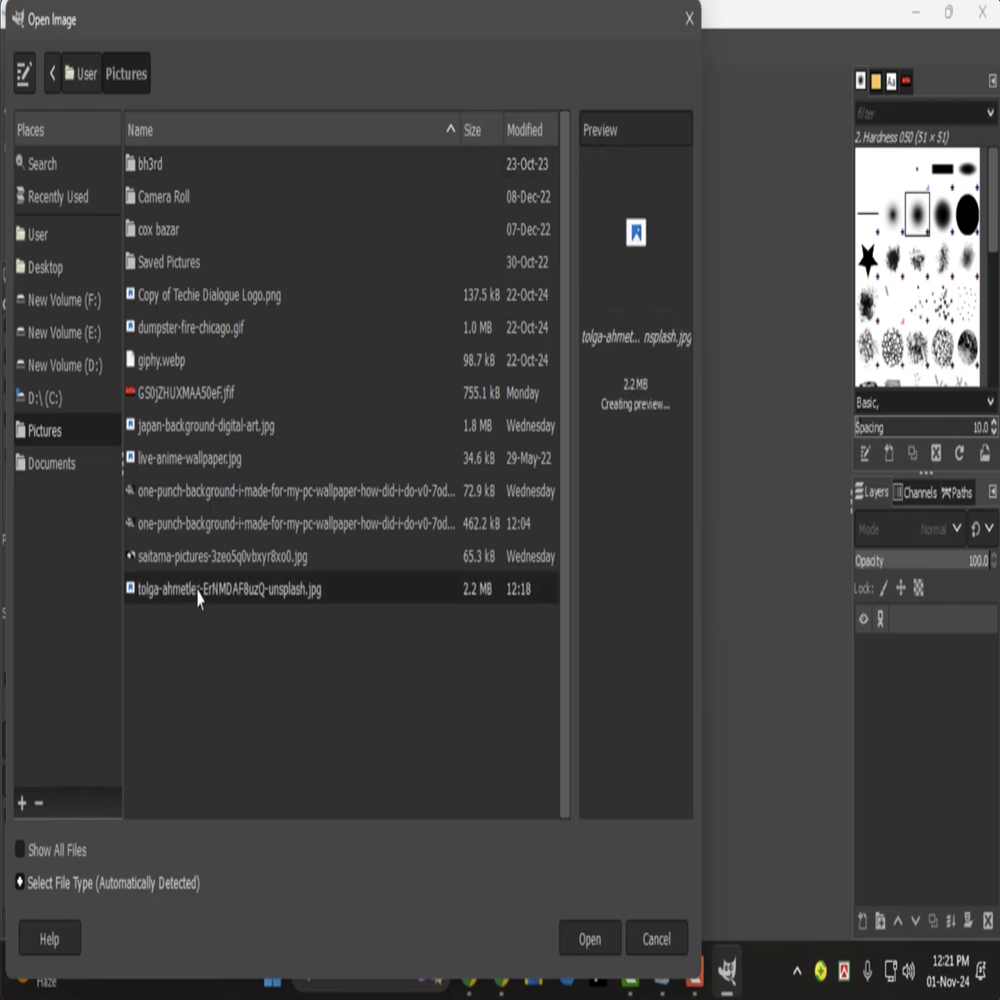} \\
\textbf{Action 6} \\
\midrule
click (585, 929) \\
\midrule
\bottomrule
\end{tabular}

\label{tab:idm_traj1_2}
\end{table*}

\begin{table*}[h]
\caption{\textbf{\ourmethod{} labeling example}}
\centering
\begin{tabular}{c}
\toprule
\textbf{Observation 7 (GIMP)} \\
\midrule
\includegraphics[width=0.5\textwidth]{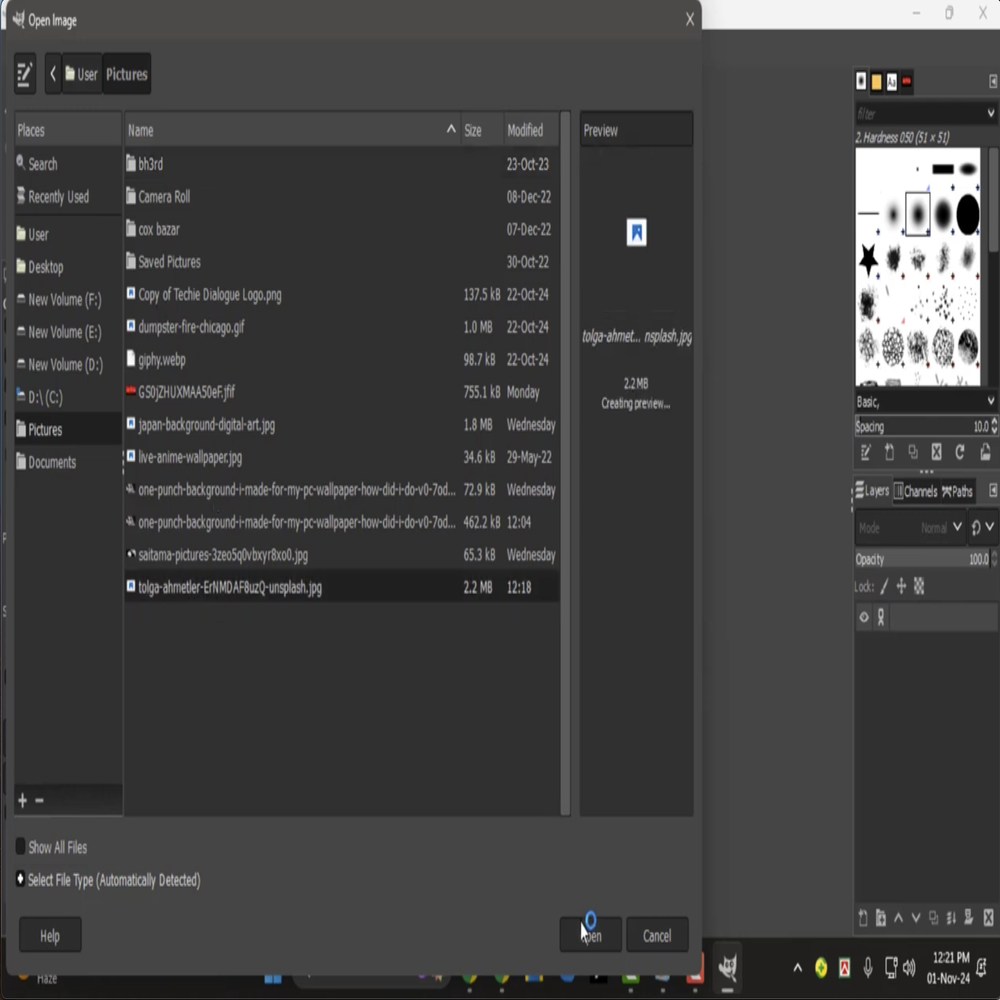} \\
\midrule
\textbf{Action 7} \\
\midrule
click (585, 929) \\
\midrule
\textbf{Observation 8 (GIMP)} \\
\midrule
\includegraphics[width=0.5\textwidth]{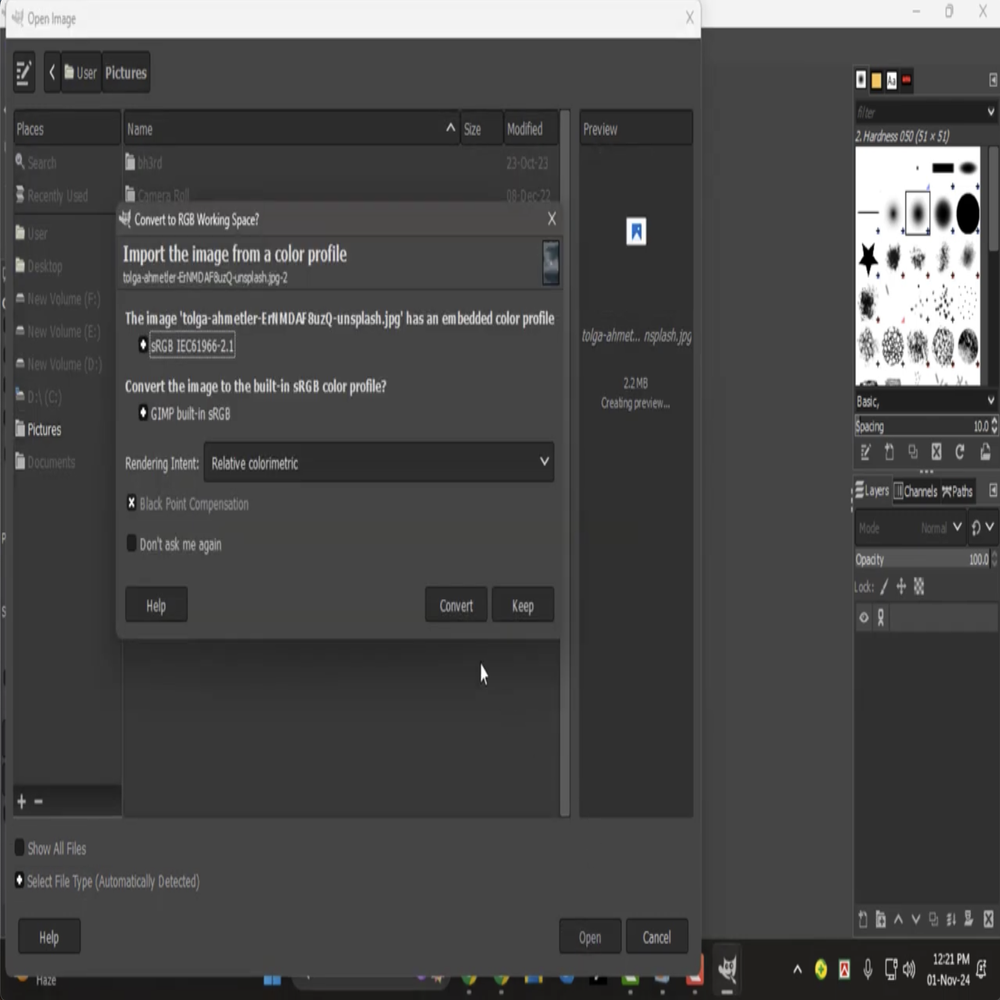} \\
\textbf{Action 8} \\
\midrule
wait \\
\midrule
\bottomrule
\end{tabular}

\label{tab:idm_traj1_3}
\end{table*}

\begin{table*}[h]
\caption{\textbf{\ourmethod{} labeling example}}
\centering
\begin{tabular}{c}
\toprule
\textbf{Observation 9 (GIMP)} \\
\midrule
\includegraphics[width=0.5\textwidth]{arxiv_google/data_samples/idm_traj1/8.png} \\
\midrule
\textbf{Action 9} \\
\midrule
move (467, 612) \\
\midrule
\textbf{Observation 10 (GIMP)} \\
\midrule
\includegraphics[width=0.5\textwidth]{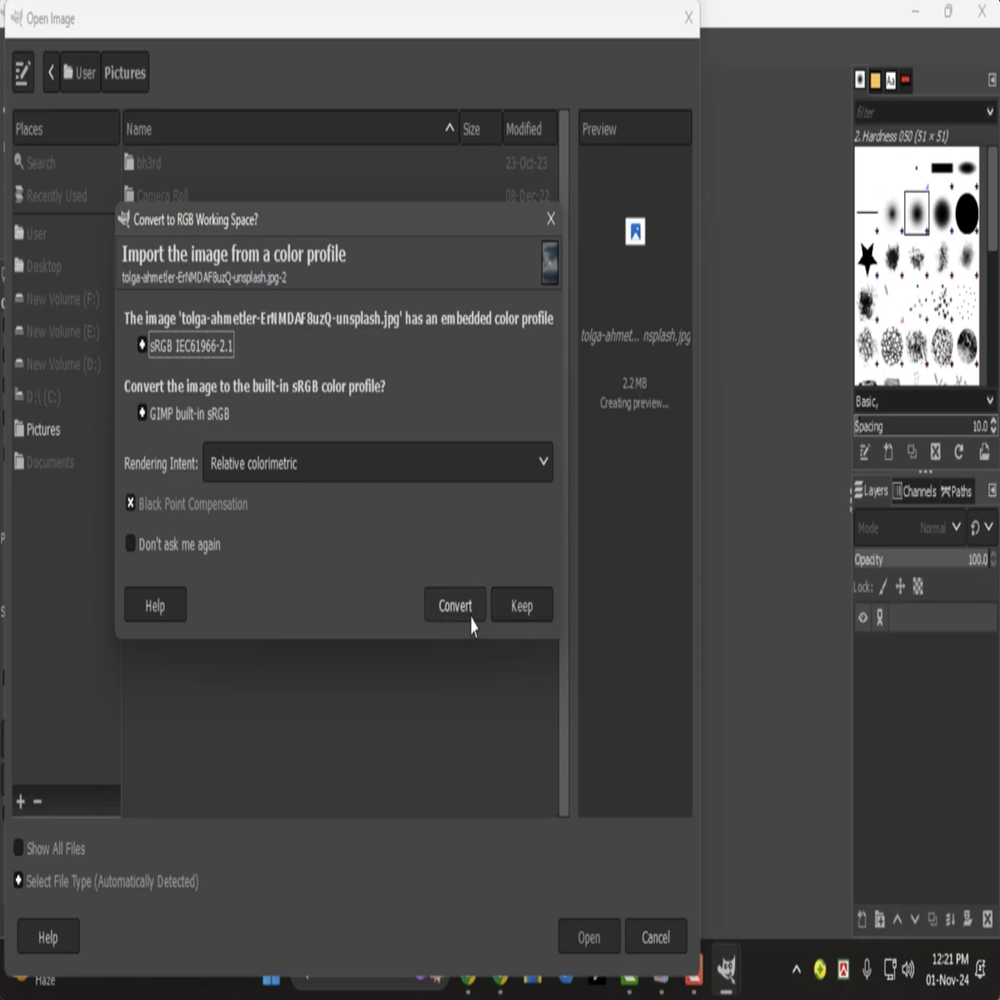} \\
\textbf{Action 10} \\
\midrule
click (467, 612) \\
\midrule
\bottomrule
\end{tabular}

\label{tab:idm_traj1_4}
\end{table*}

\begin{table*}[h]
\caption{\textbf{\ourmethod{} labeling example}}
\centering
\begin{tabular}{c}
\toprule
\textbf{Observation 11 (GIMP)} \\
\midrule
\includegraphics[width=0.5\textwidth]{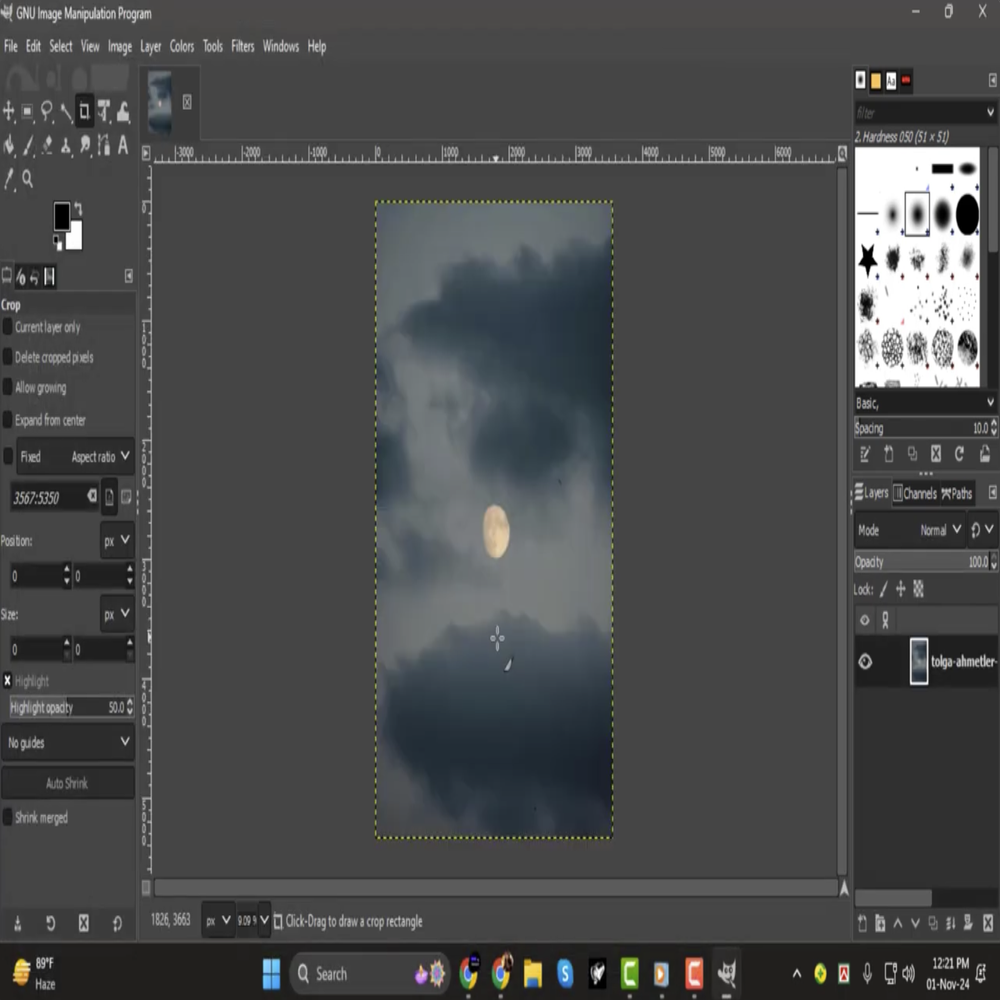} \\
\midrule
\textbf{Action 11} \\
\midrule
click (121, 46) \\
\midrule
\textbf{Observation 12 (GIMP)} \\
\midrule
\includegraphics[width=0.5\textwidth]{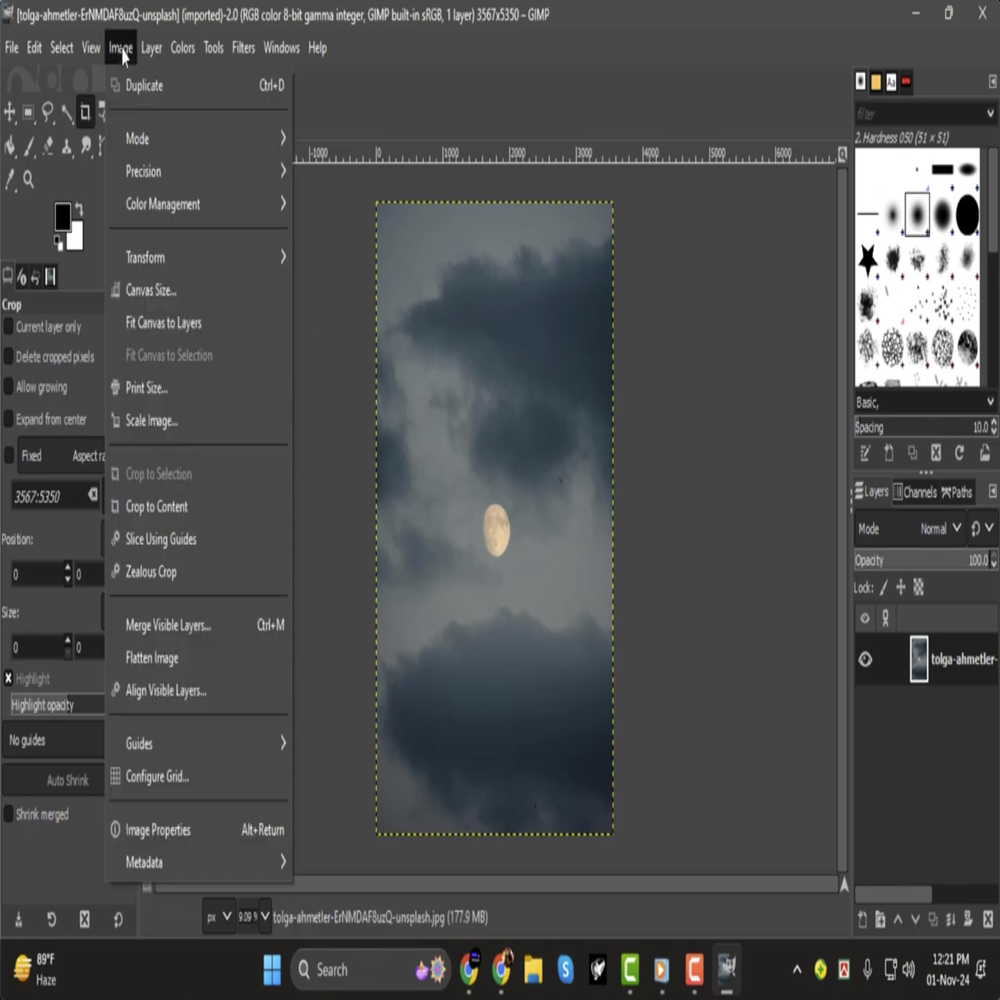} \\
\textbf{Action 12} \\
\midrule
click (145, 426) \\
\midrule
\bottomrule
\end{tabular}

\label{tab:idm_traj1_5}
\end{table*}

\begin{table*}[h]
\caption{\textbf{\ourmethod{} labeling example}}
\centering
\begin{tabular}{c}
\toprule
\textbf{Observation 13 (GIMP)} \\
\midrule
\includegraphics[width=0.5\textwidth]{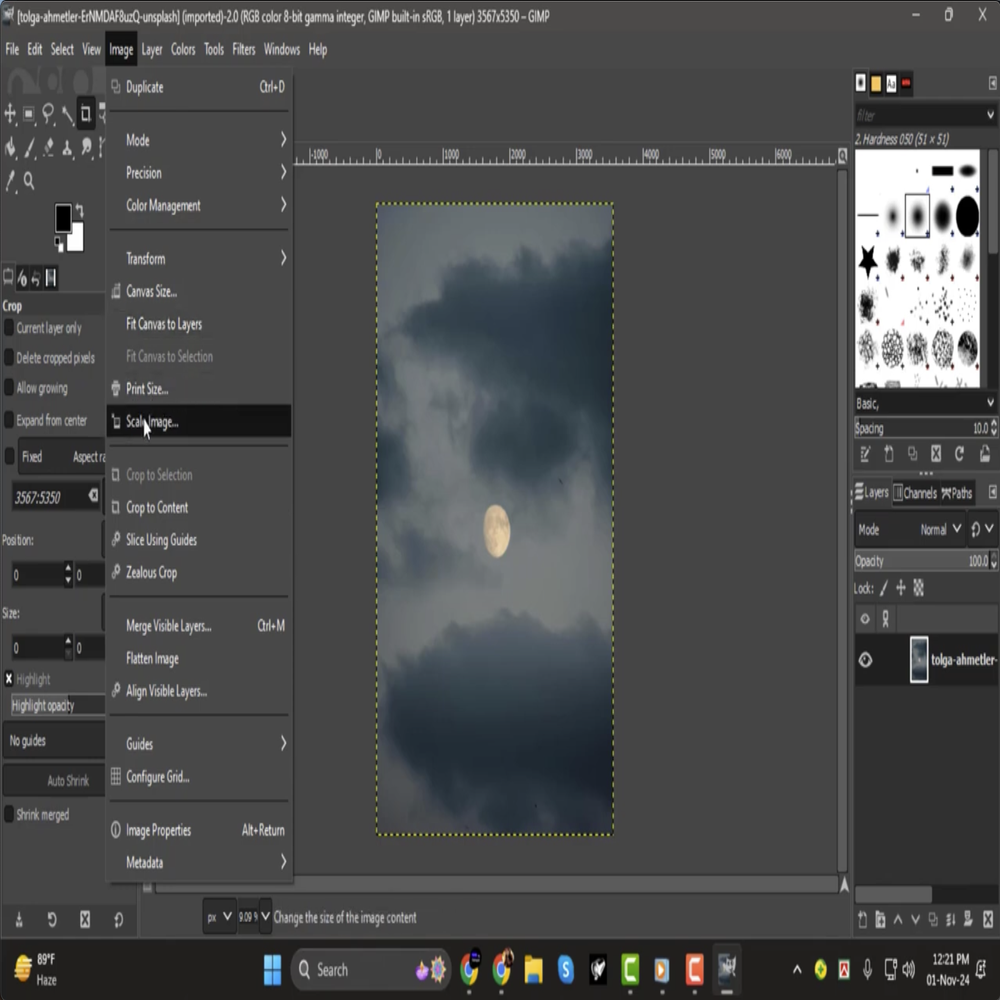} \\
\midrule
\textbf{Action 13} \\
\midrule
click (459, 411) \\
\midrule
\textbf{Observation 14 (GIMP)} \\
\midrule
\includegraphics[width=0.5\textwidth]{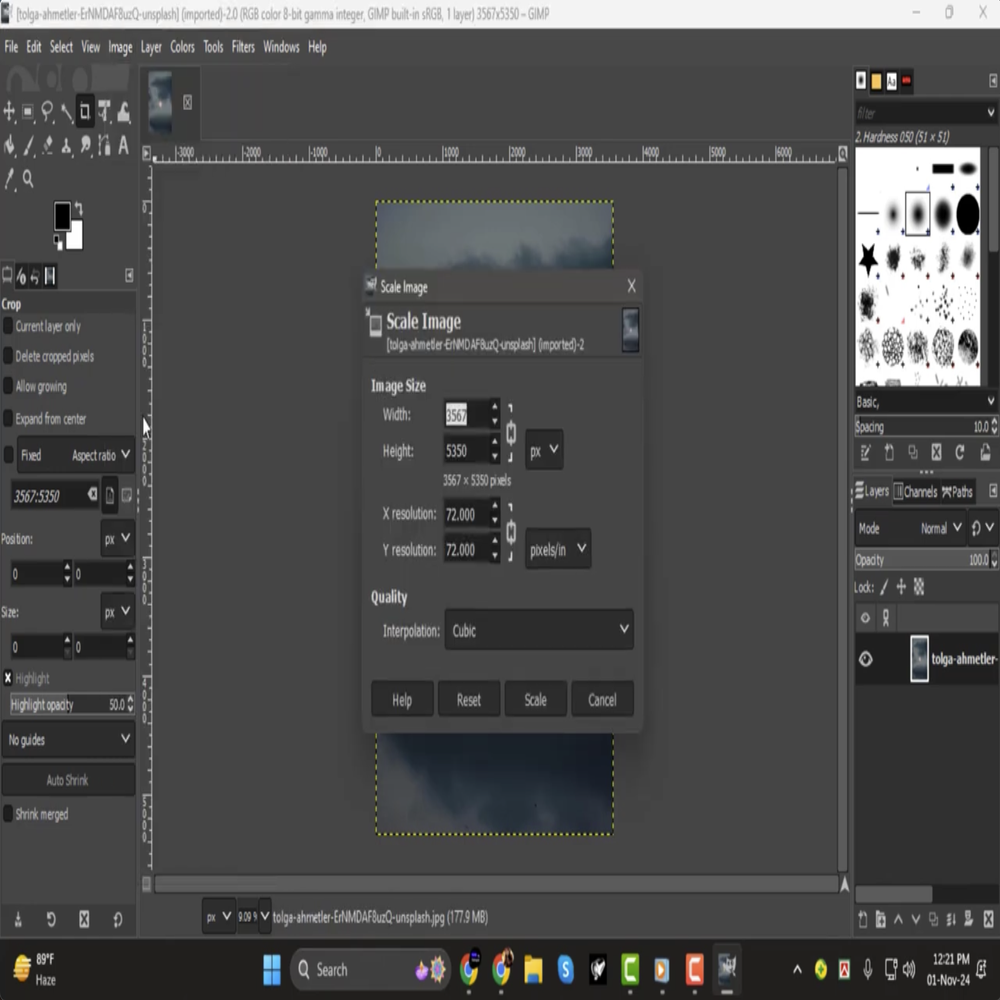} \\
\textbf{Action 14} \\
\midrule
type (458, 412, "5000") \\
\midrule
\bottomrule
\end{tabular}

\label{tab:idm_traj1_6}
\end{table*}

\begin{table*}[h]
\caption{\textbf{\ourmethod{} labeling example}}
\centering
\begin{tabular}{c}
\toprule
\textbf{Observation 15 (GIMP)} \\
\midrule
\includegraphics[width=0.5\textwidth]{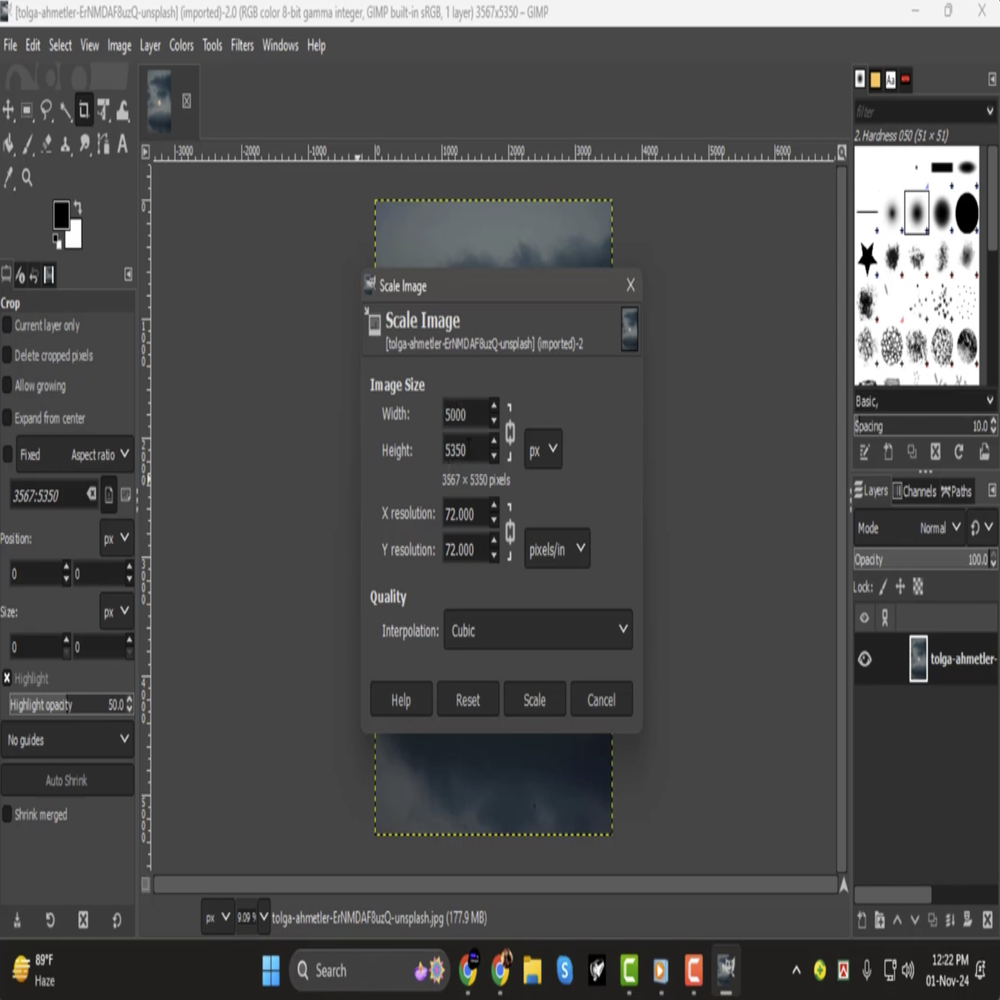} \\
\midrule
\textbf{Action 15} \\
\midrule
type (454, 446, "7499") \\
\midrule
\textbf{Observation 16 (GIMP)} \\
\midrule
\includegraphics[width=0.5\textwidth]{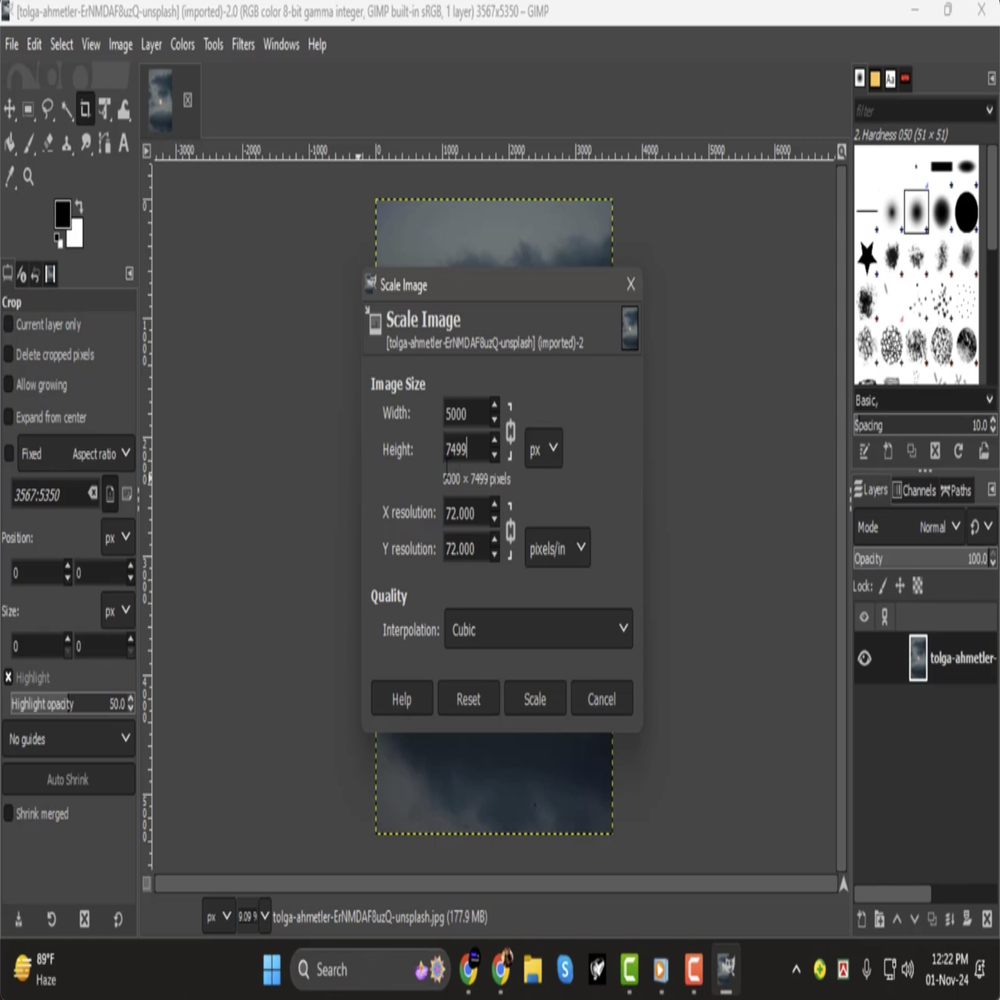} \\
\textbf{Action 16} \\
\midrule
click (534, 704) \\
\midrule
\bottomrule
\end{tabular}

\label{tab:idm_traj1_7}
\end{table*}

\begin{table*}[h]
\caption{\textbf{\ourmethod{} labeling example}}
\centering
\begin{tabular}{c}
\toprule
\textbf{Observation 17 (GIMP)} \\
\midrule
\includegraphics[width=0.5\textwidth]{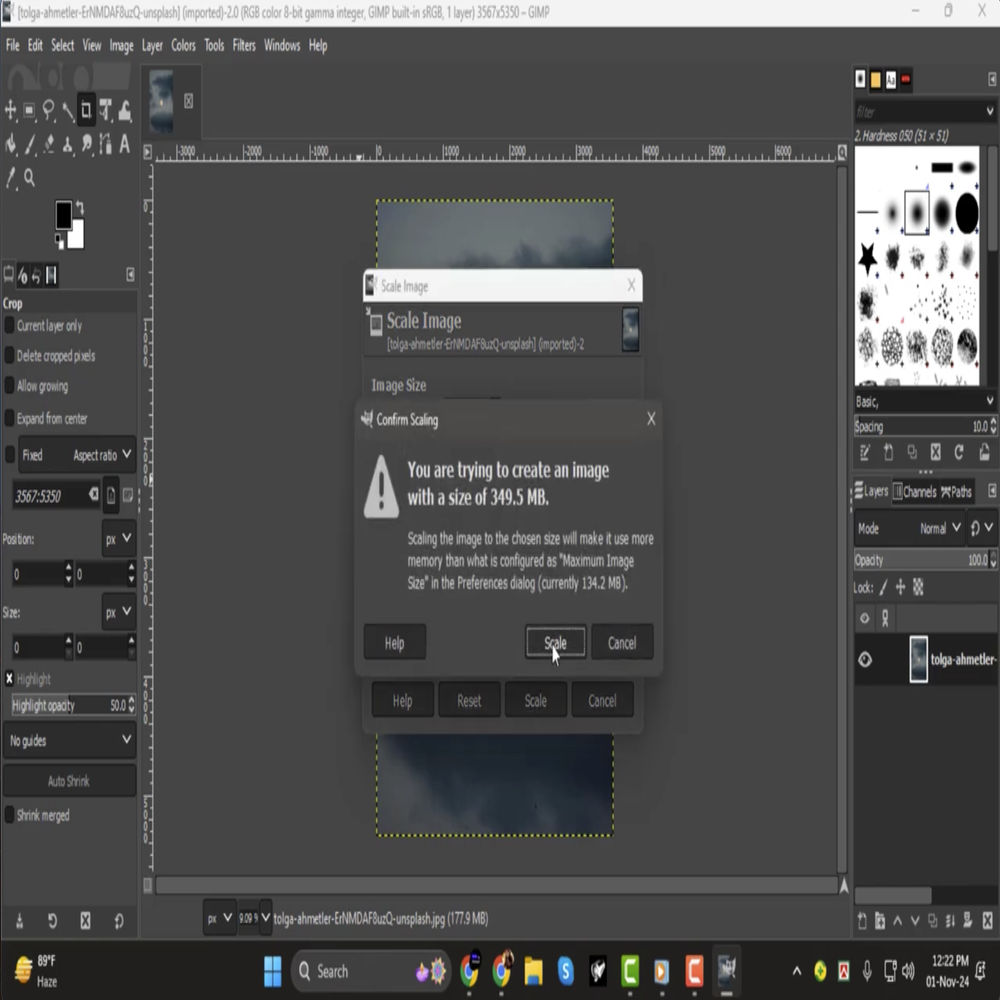} \\
\midrule
\textbf{Action 17} \\
\midrule
click (555, 647) \\
\midrule
\textbf{Observation 18 (GIMP)} \\
\midrule
\includegraphics[width=0.5\textwidth]{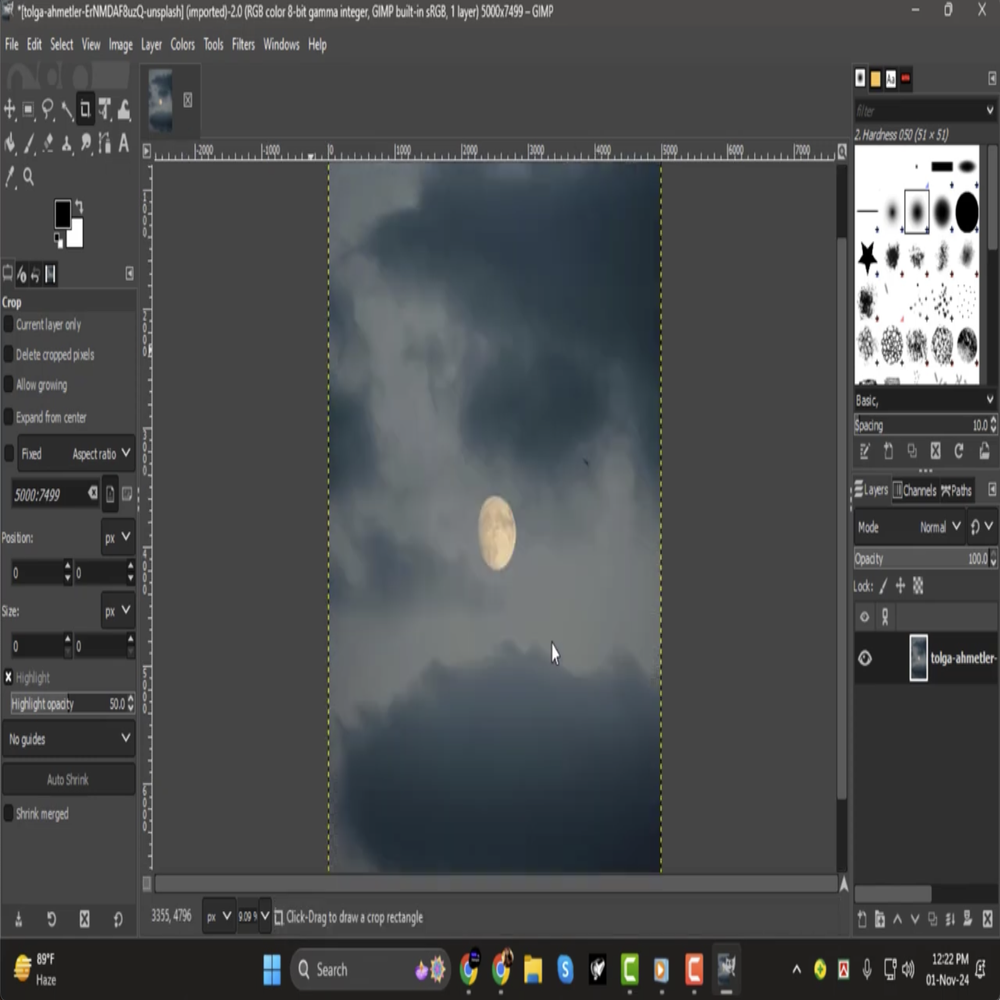} \\
\textbf{Action 18} \\
\midrule
No action (Done) \\
\midrule
\bottomrule
\end{tabular}

\label{tab:idm_traj1_8}
\end{table*}


\begin{table*}[h]
\caption{\textbf{OSWorld example (UI-TARS-1.5-7B w/~\ourmethod})}
\centering
\begin{tabular}{c}
\toprule
Task: Current volume is 125\%. Can you increase it to 200\% in VLC? \\
\midrule
\textbf{Observation 1 (VLC)} \\
\midrule
\includegraphics[width=0.7\textwidth]{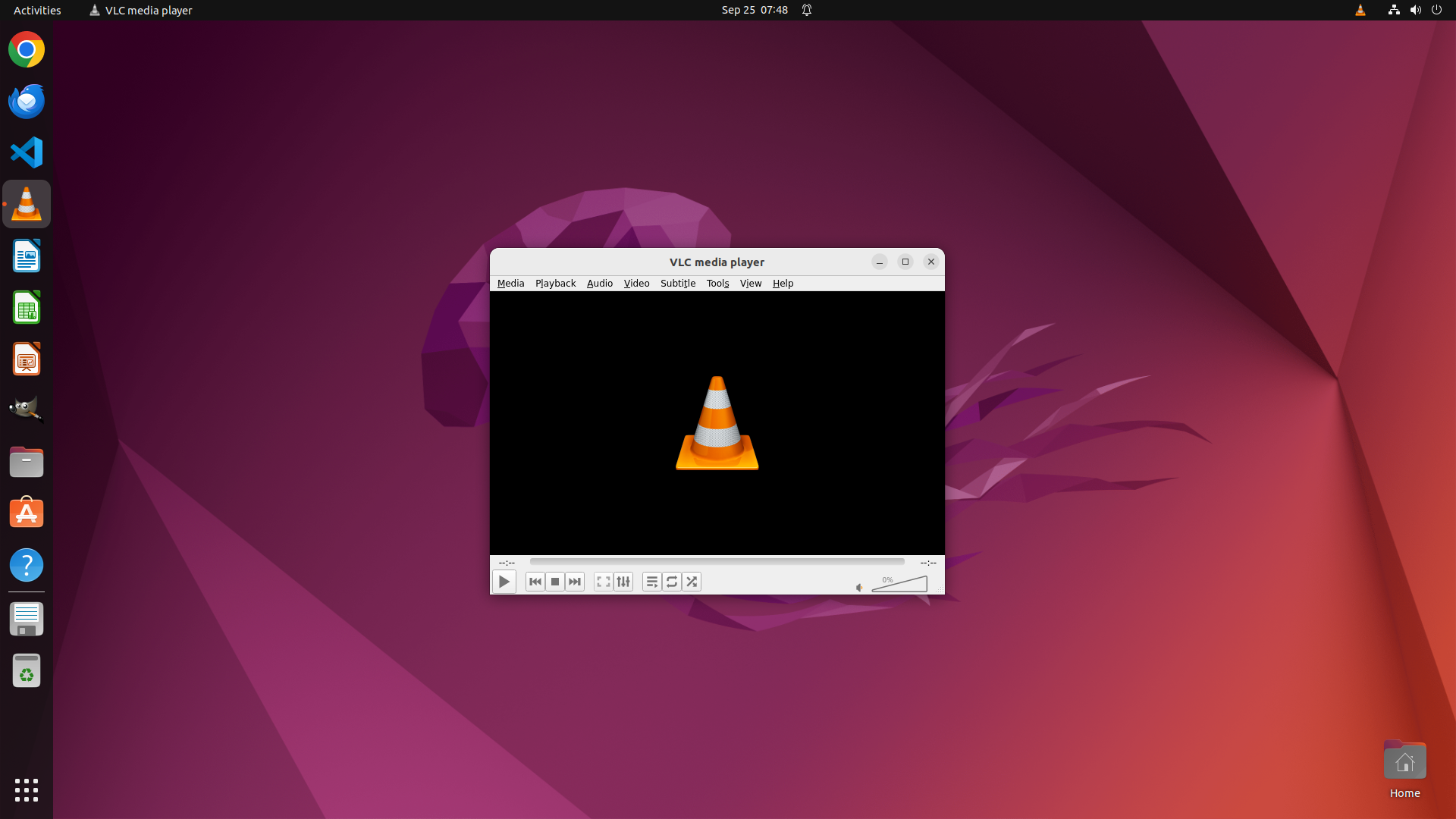} \\
\midrule
\textbf{Action 1} \\
\midrule
import pyautogui \\
pyautogui.click(941, 369) \\
\midrule
\textbf{Observation 2 (VLC)} \\
\midrule
\includegraphics[width=0.7\textwidth]{arxiv_google/data_samples/traj4/Ubuntu0-2025-09-24-19-48-50.png} \\
\textbf{Action 2} \\
\midrule
import pyautogui \\
pyautogui.click(984, 602) \\
\midrule
\bottomrule
\end{tabular}

\label{tab:osworld_traj1_0}
\end{table*}

\begin{table*}[h]
\caption{\textbf{OSWorld example (UI-TARS-1.5-7B w/~\ourmethod})}
\centering
\begin{tabular}{c}
\toprule
\textbf{Observation 3 (VLC)} \\
\midrule
\includegraphics[width=0.7\textwidth]{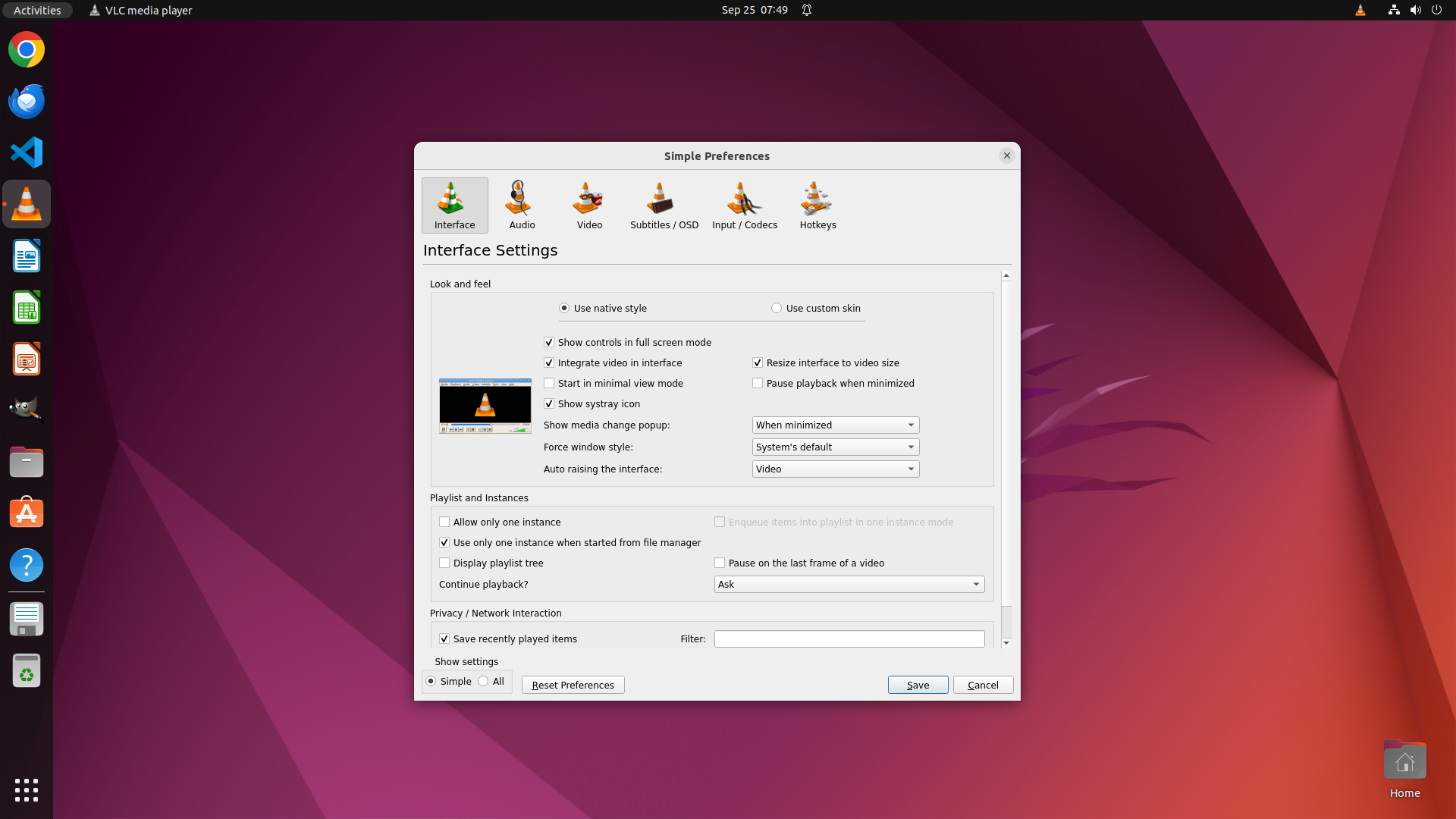} \\
\midrule
\textbf{Action 3} \\
\midrule
import pyautogui \\
pyautogui.click(635, 892) \\
\midrule
\textbf{Observation 4 (VLC)} \\
\midrule
\includegraphics[width=0.7\textwidth]{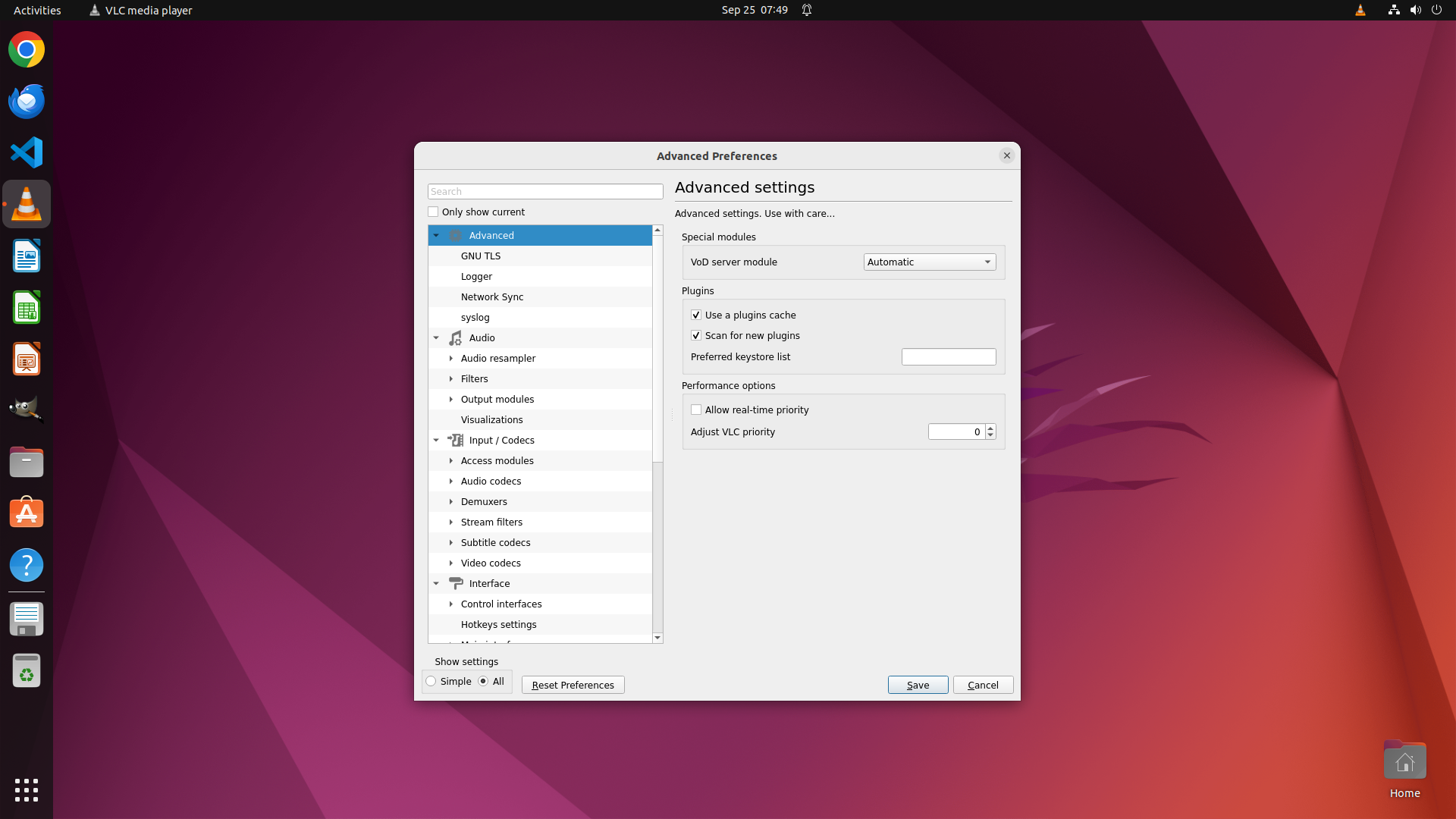} \\
\textbf{Action 4} \\
\midrule
import pyautogui \\
pyautogui.click(642, 763, button="left") \\
\midrule
\bottomrule
\end{tabular}

\label{tab:osworld_traj1_1}
\end{table*}

\begin{table*}[h]
\caption{\textbf{OSWorld example (UI-TARS-1.5-7B w/~\ourmethod})}
\centering
\begin{tabular}{c}
\toprule
\textbf{Observation 5 (VLC)} \\
\midrule
\includegraphics[width=0.7\textwidth]{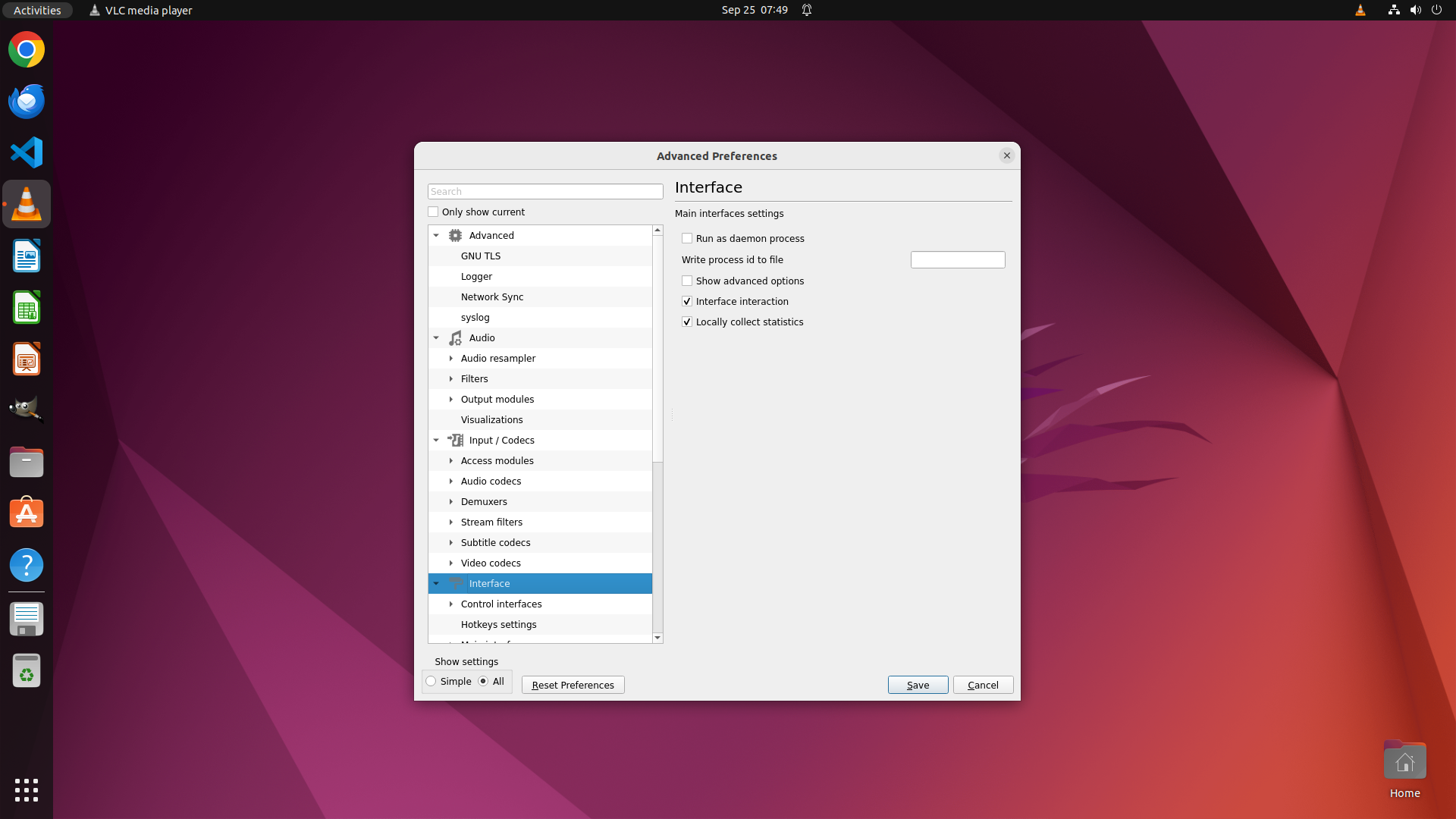} \\
\midrule
\textbf{Action 5} \\
\midrule
import pyautogui \\
pyautogui.scroll(-50, x=722, y=596) \\
\midrule
\textbf{Observation 6 (VLC)} \\
\midrule
\includegraphics[width=0.7\textwidth]{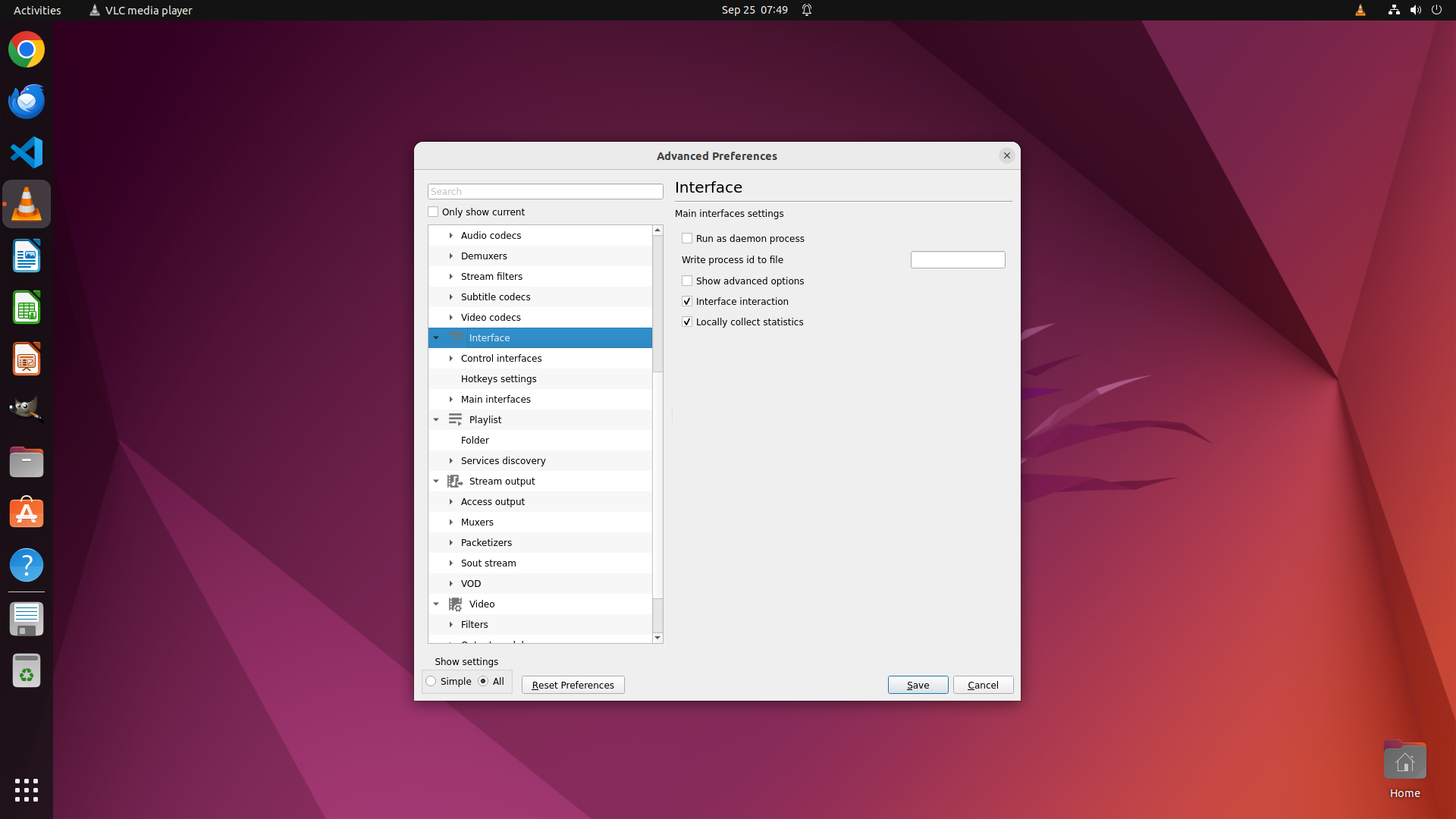} \\
\textbf{Action 6} \\
\midrule
import pyautogui \\
pyautogui.click(614, 518) \\
\midrule
\bottomrule
\end{tabular}

\label{tab:osworld_traj1_2}
\end{table*}

\begin{table*}[h]
\caption{\textbf{OSWorld example (UI-TARS-1.5-7B w/~\ourmethod})}
\centering
\begin{tabular}{c}
\toprule
\textbf{Observation 7 (VLC)} \\
\midrule
\includegraphics[width=0.7\textwidth]{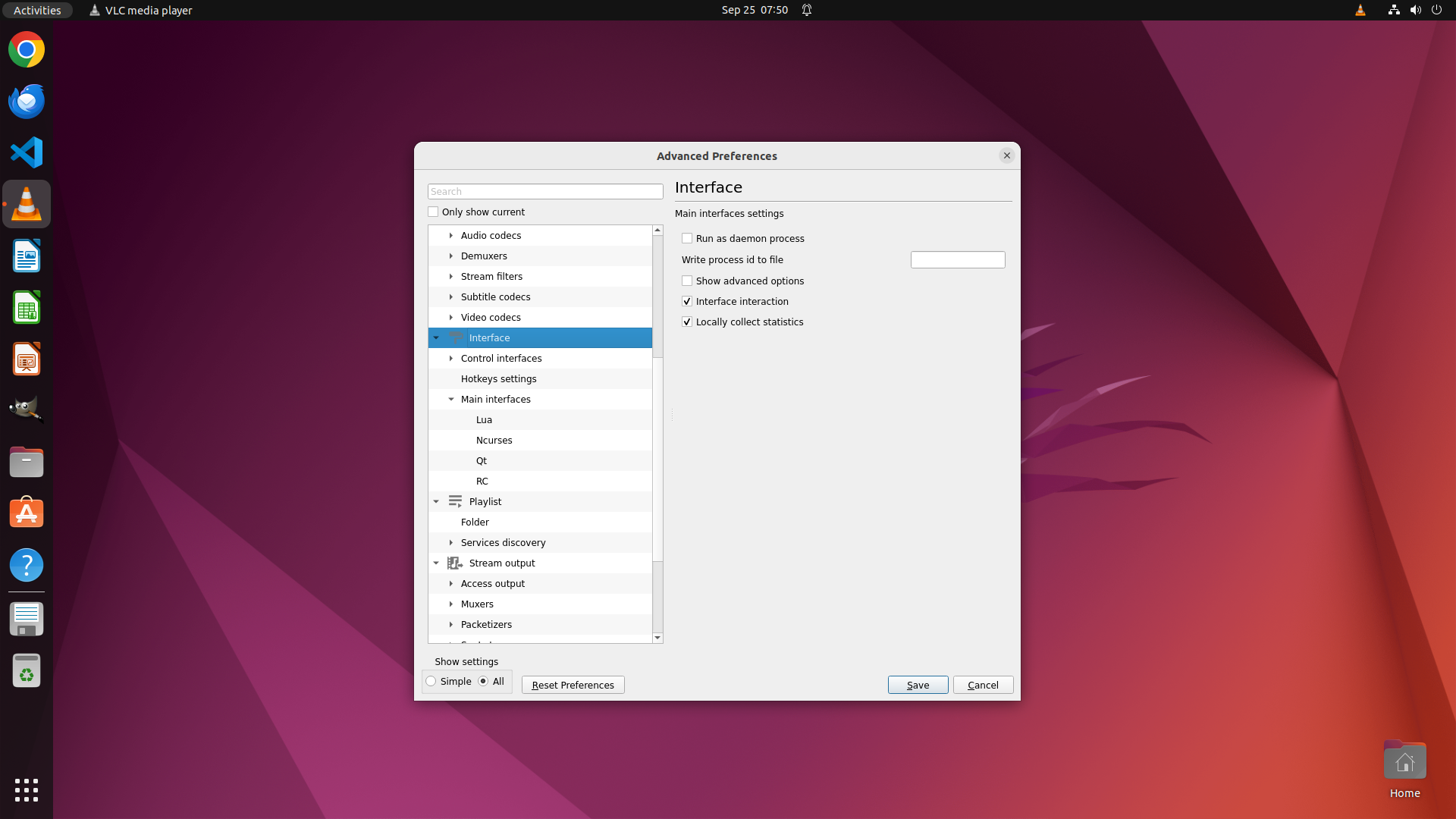} \\
\midrule
\textbf{Action 7} \\
\midrule
import pyautogui \\
pyautogui.click(675, 604) \\
\midrule
\textbf{Observation 8 (VLC)} \\
\midrule
\includegraphics[width=0.7\textwidth]{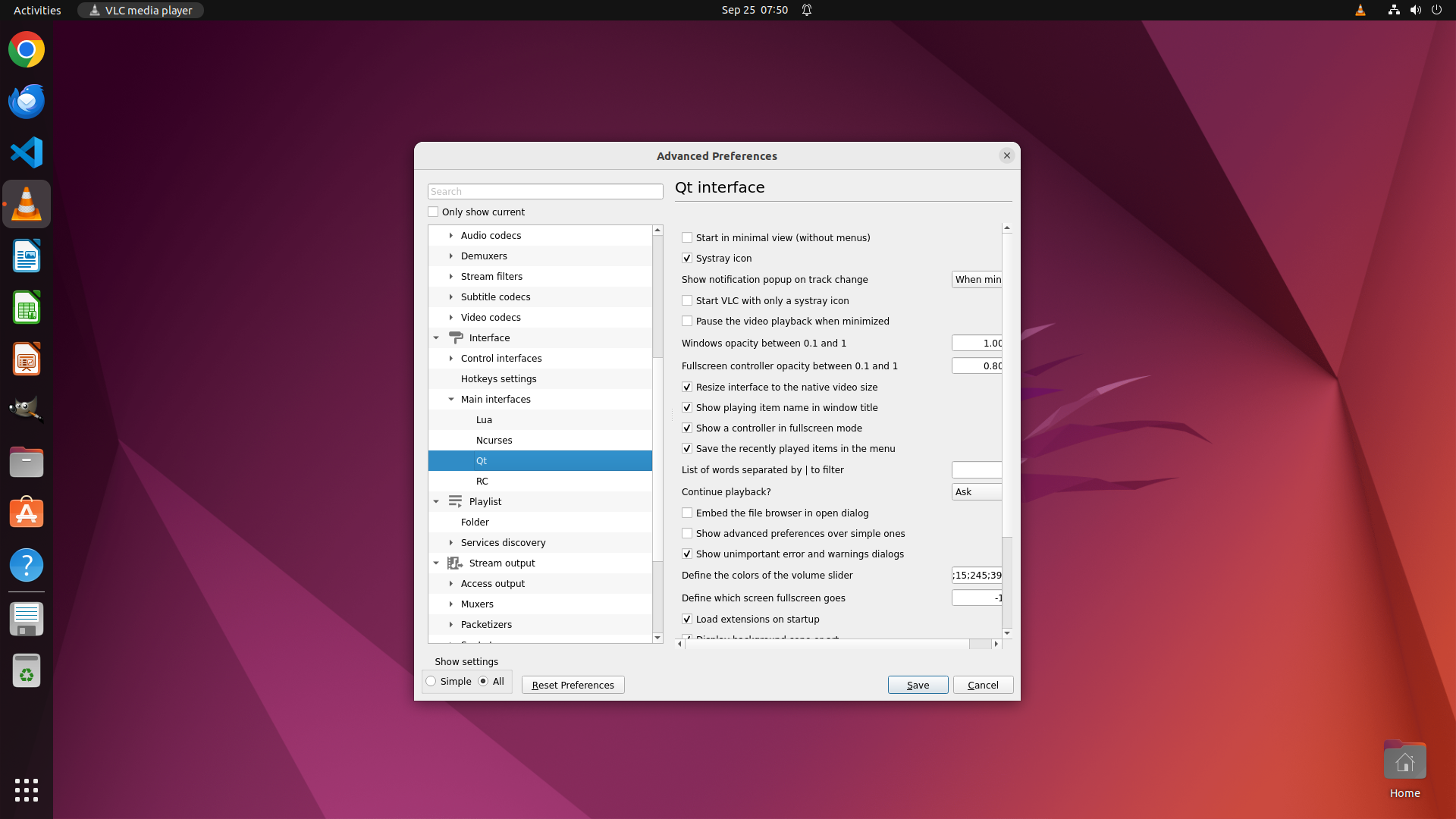} \\
\textbf{Action 8} \\
\midrule
import pyautogui \\
pyautogui.scroll(-50, x=1115, y=583) \\
\midrule
\bottomrule
\end{tabular}

\label{tab:osworld_traj1_3}
\end{table*}

\begin{table*}[h]
\caption{\textbf{OSWorld example (UI-TARS-1.5-7B w/~\ourmethod})}
\centering
\begin{tabular}{c}
\toprule
\textbf{Observation 9 (VLC)} \\
\midrule
\includegraphics[width=0.7\textwidth]{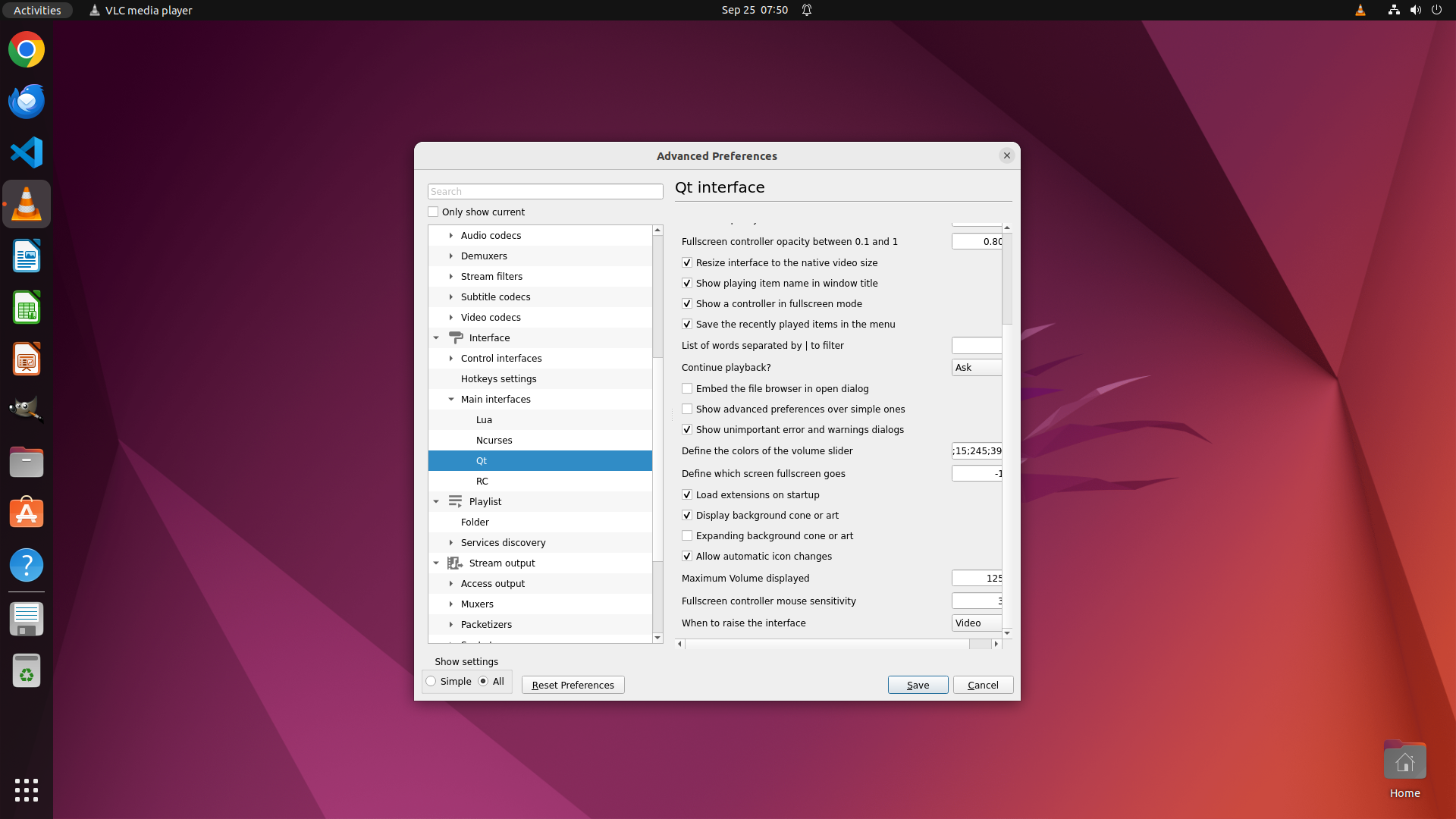} \\
\midrule
\textbf{Action 9} \\
\midrule
import pyautogui \\
pyautogui.click(1292, 763, button="left") \\
\midrule
\textbf{Observation 10 (VLC)} \\
\midrule
\includegraphics[width=0.7\textwidth]{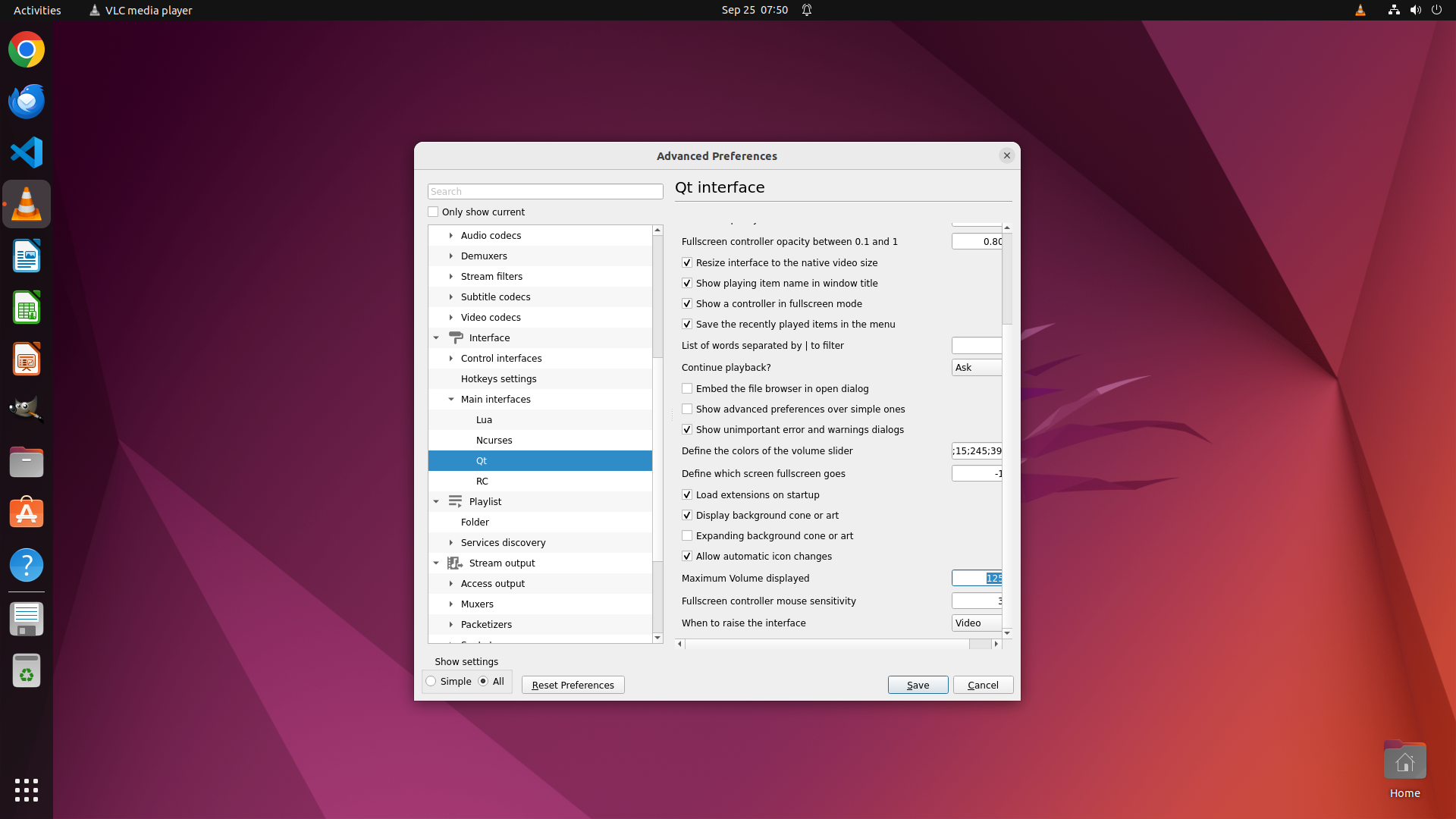} \\
\textbf{Action 10} \\
\midrule
import pyautogui \\
pyautogui.type("300") \\
\midrule
\bottomrule
\end{tabular}

\label{tab:osworld_traj1_4}
\end{table*}

\begin{table*}[h]
\caption{\textbf{OSWorld example (UI-TARS-1.5-7B w/~\ourmethod})}
\centering
\begin{tabular}{c}
\toprule
\textbf{Observation 11 (VLC)} \\
\midrule
\includegraphics[width=0.7\textwidth]{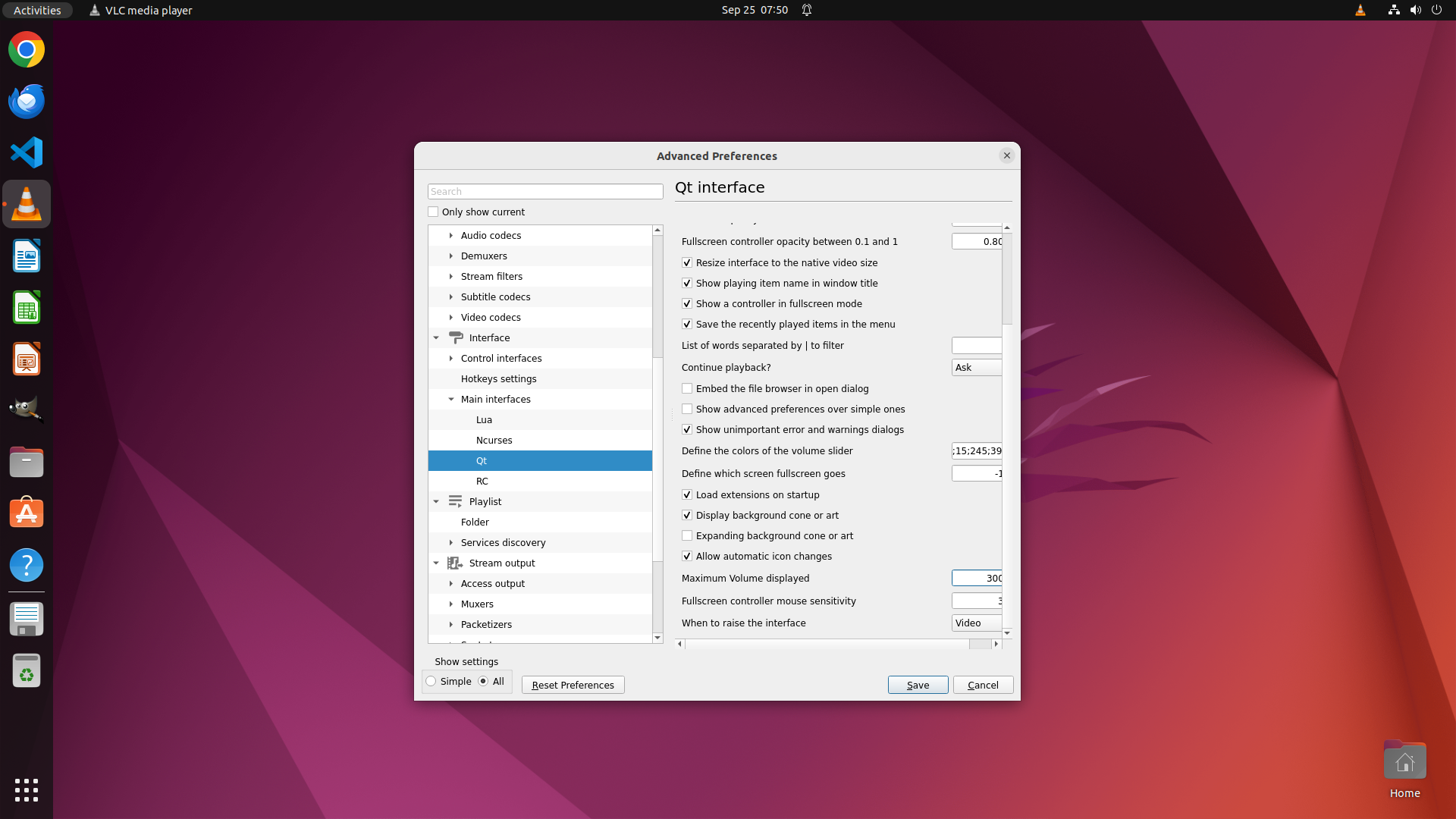} \\
\midrule
\textbf{Action 11} \\
\midrule
import pyautogui \\
pyautogui.click(1205, 902) \\
\midrule
\textbf{Observation 12 (VLC)} \\
\midrule
\includegraphics[width=0.7\textwidth]{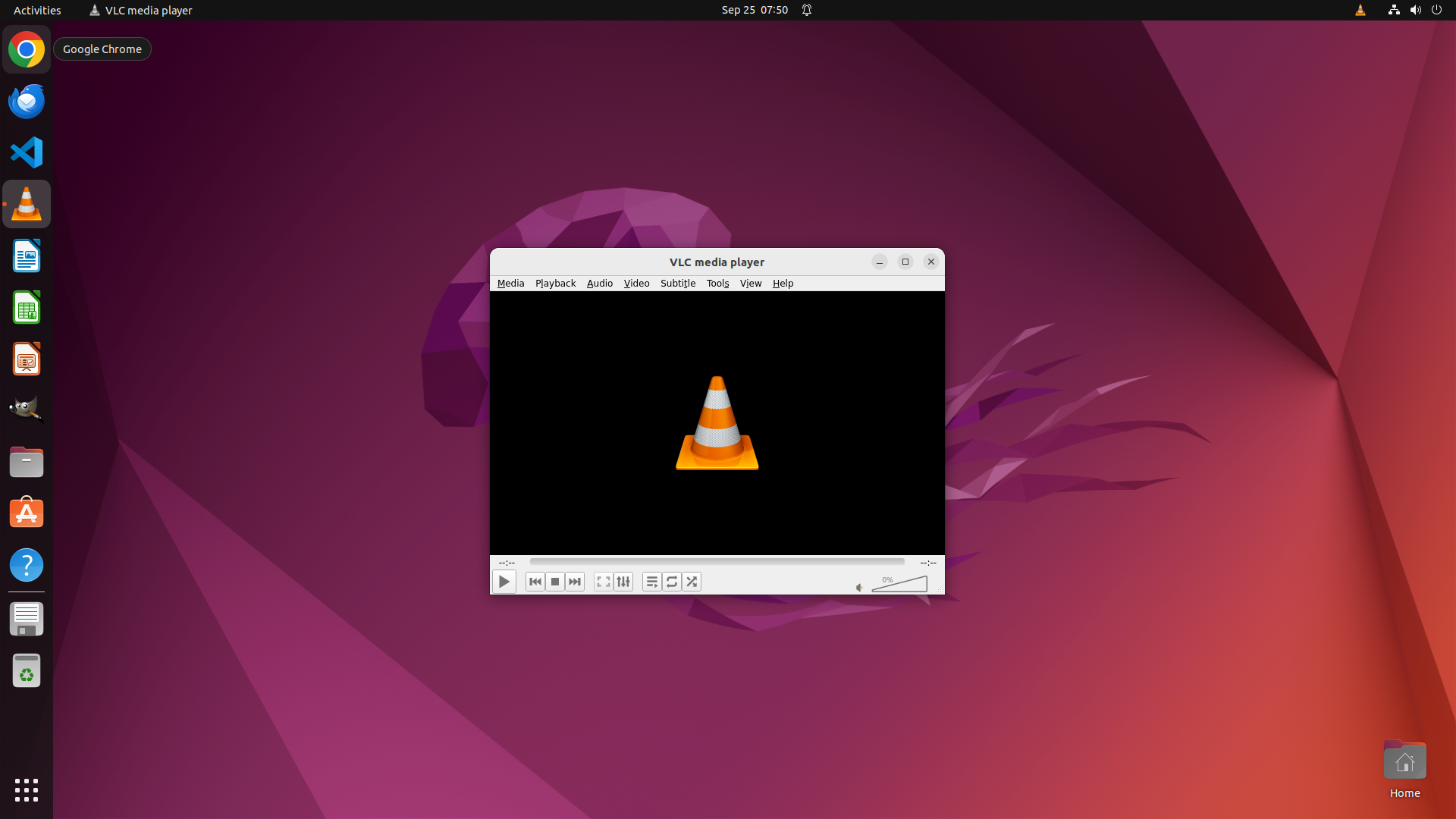} \\
\textbf{Action 12} \\
\midrule
"DONE" \\
\midrule
\bottomrule
\end{tabular}

\label{tab:osworld_traj1_5}
\end{table*}


\begin{table*}[h]
\caption{\textbf{WindowsAgentArena example (UI-TARS-1.5-7B w/~\ourmethod})}
\centering
\begin{tabular}{c}
\toprule
Task: Set the file "secret.txt" in the Documents folder as hidden. \\
\midrule
\textbf{Observation 1 (Windows File Explorer)} \\
\midrule
\includegraphics[width=0.7\textwidth]{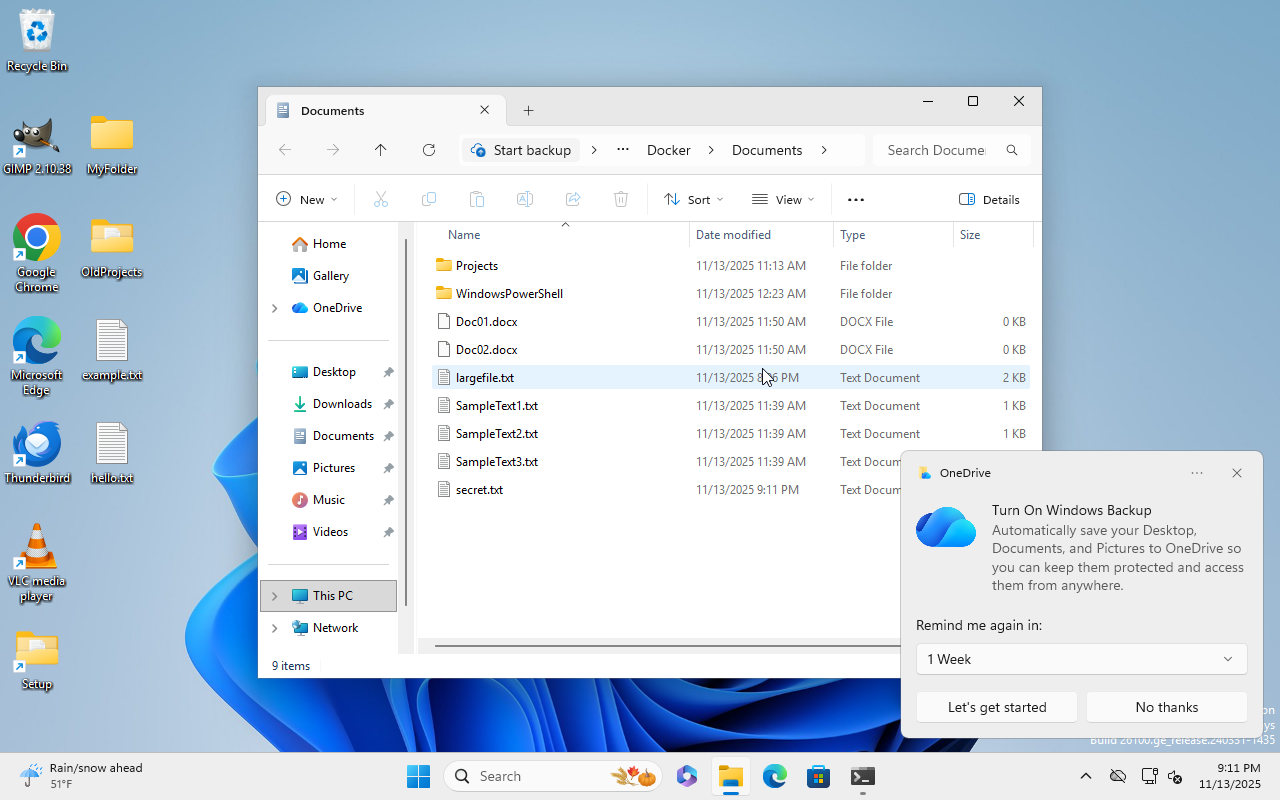} \\
\midrule
\textbf{Action 1} \\
\midrule
import pyautogui \\
pyautogui.moveTo(487, 497) \\
\midrule
\textbf{Observation 2 (Windows File Explorer)} \\
\midrule
\includegraphics[width=0.7\textwidth]{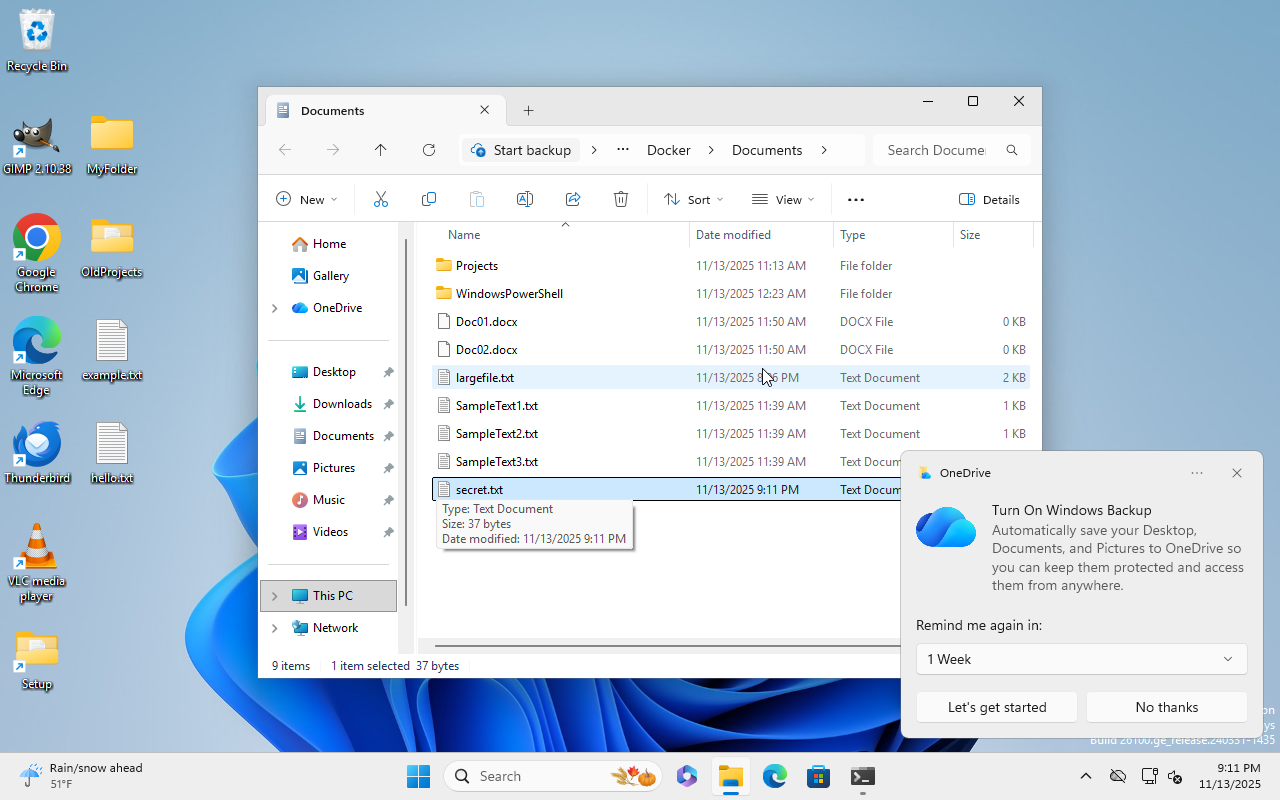} \\
\textbf{Action 2} \\
\midrule
import pyautogui \\
pyautogui.click(button='right) \\
\midrule
\bottomrule
\end{tabular}

\label{tab:windows_traj1_0}
\end{table*}

\begin{table*}[ht]
\caption{\textbf{WindowsAgentArena example (UI-TARS-1.5-7B w/~\ourmethod), cont.}}
\centering
\begin{tabular}{c}
\toprule
\textbf{Observation 3 (Windows File Explorer)} \\
\midrule
\includegraphics[width=0.7\textwidth]{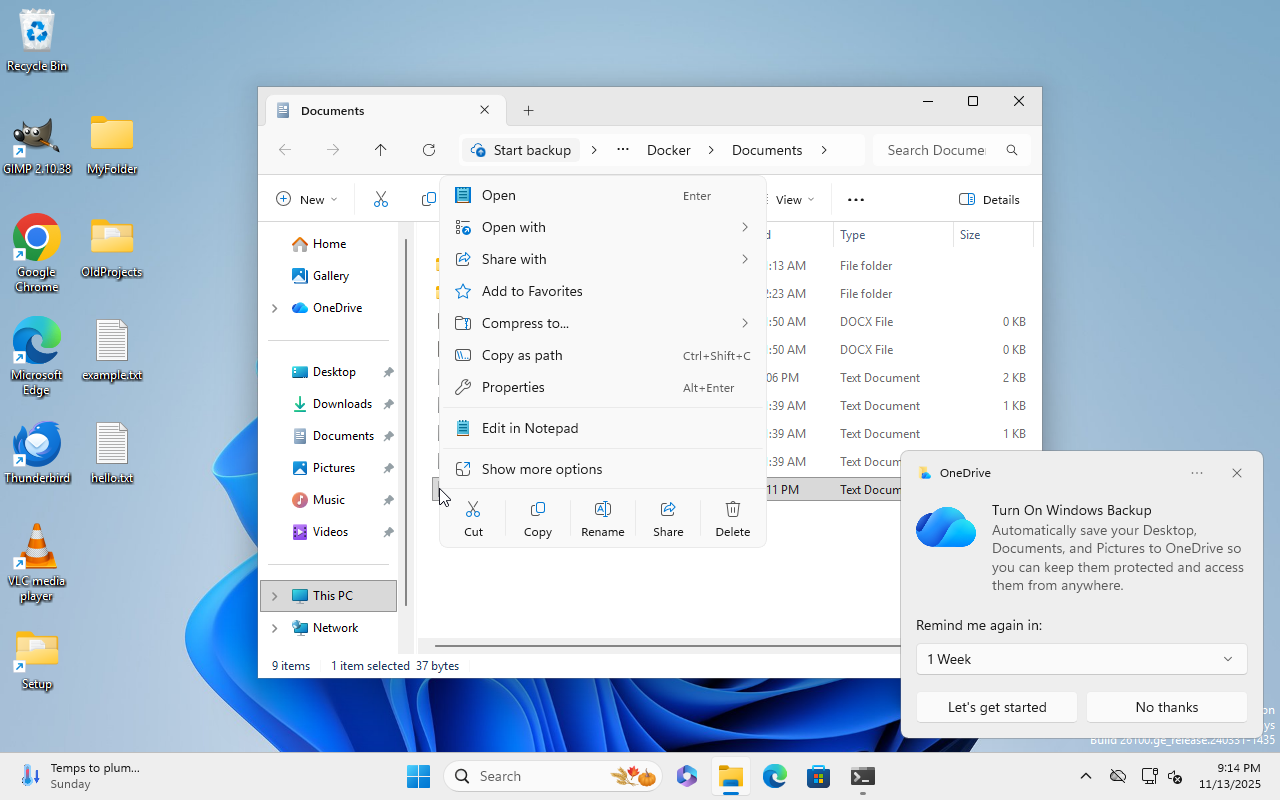} \\
\midrule
\textbf{Action 3} \\
\midrule
import pyautogui \\
pyautogui.click(500, 388) \\
\midrule
\textbf{Observation 4 (Windows File Explorer)} \\
\midrule
\includegraphics[width=0.7\textwidth]{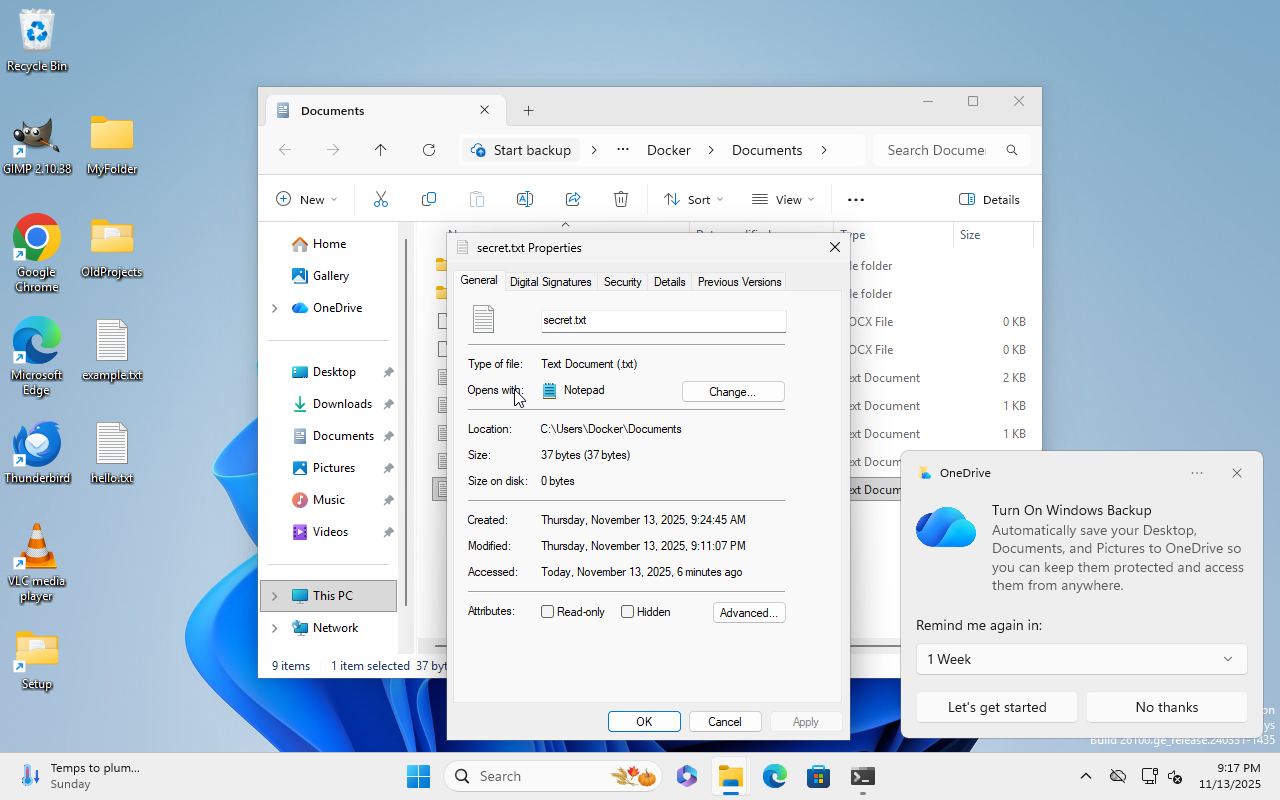} \\
\textbf{Action 4} \\
\midrule
import pyautogui \\
pyautogui.click(626, 611)  \\
\midrule
\bottomrule
\end{tabular}

\label{tab:windows_traj1_1}
\end{table*}

\begin{table*}[ht]
\caption{\textbf{WindowsAgentArena example (UI-TARS-1.5-7B w/~\ourmethod), cont.}}
\centering
\begin{tabular}{c}
\toprule
\textbf{Observation 5 (Windows File Explorer)} \\
\midrule
\includegraphics[width=0.7\textwidth]{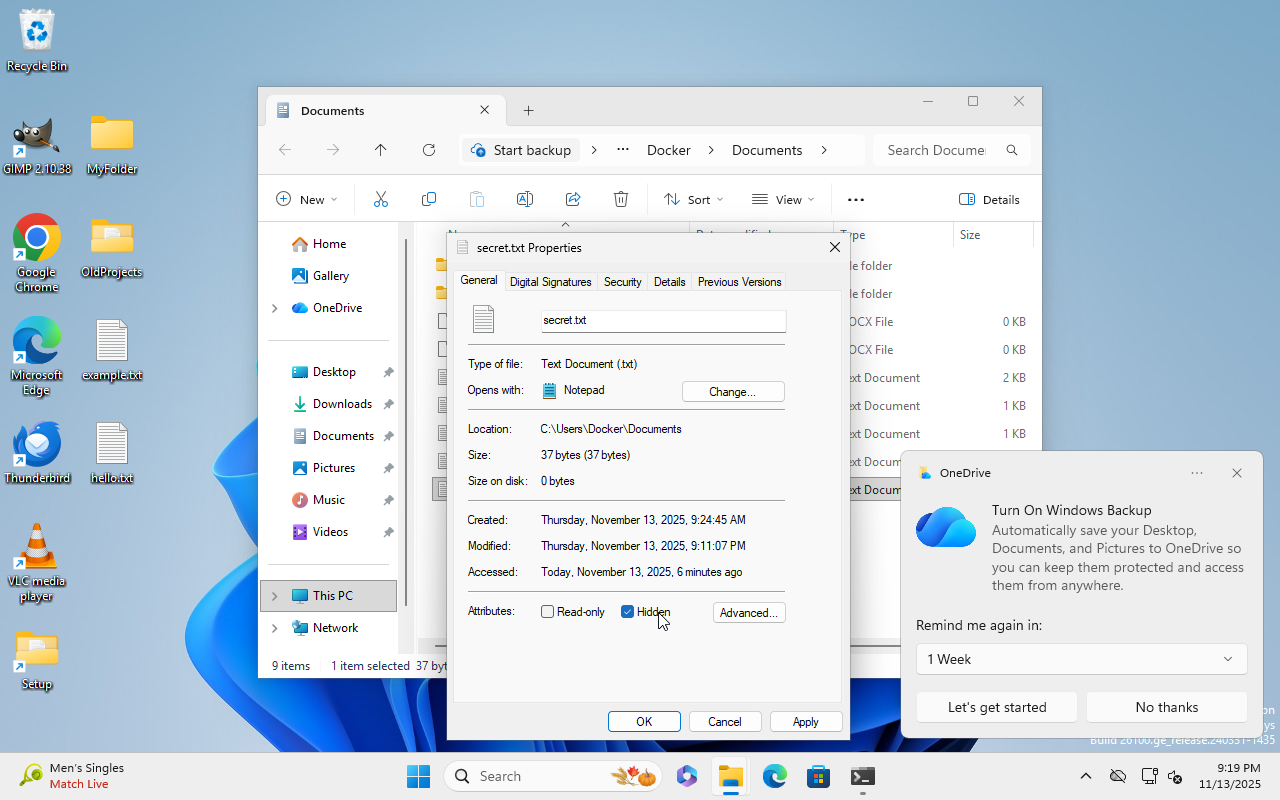} \\
\midrule
\textbf{Action 5} \\
\midrule
import pyautogui \\
pyautogui.click(635, 719) \\
\midrule
\textbf{Observation 6 (Windows File Explorer)} \\
\midrule
\includegraphics[width=0.7\textwidth]{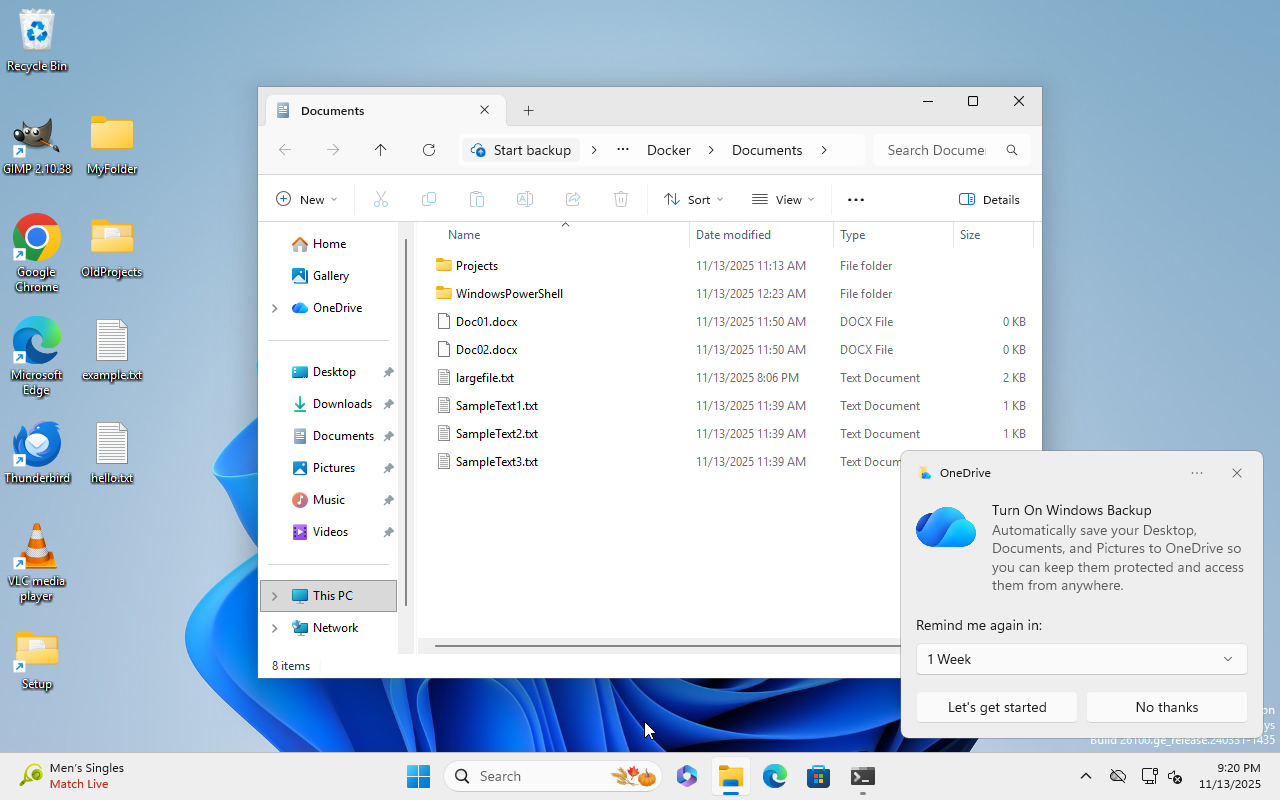} \\
\textbf{Action 6} \\
\midrule
"DONE" \\
\midrule
\bottomrule
\end{tabular}

\label{tab:windows_traj1_2}
\end{table*}

\end{document}